\pdfoutput=1

\documentclass[11pt]{article}

\usepackage[final]{acl}

\usepackage{times}
\usepackage{latexsym}

\usepackage[T1]{fontenc}

\usepackage[utf8]{inputenc}

\usepackage{microtype}

\usepackage{inconsolata}

\usepackage{graphicx}

\usepackage{bbding}
\definecolor{green}{HTML}{009B55}
\usepackage{microtype}
\usepackage{graphicx}
\usepackage{subfigure}
\usepackage{threeparttable}
\usepackage{booktabs} 
\usepackage{xspace}
\usepackage[ruled,vlined]{algorithm2e}
\usepackage{macros}
\usepackage{wrapfig}
\usepackage{lipsum}
\usepackage{amsfonts}       
\usepackage{nicefrac}       
\usepackage{xcolor, colortbl}         
\usepackage{multicol}
\usepackage{multirow}
\usepackage{enumitem}

\usepackage{listings}
\usepackage{pythonhighlight}

\usepackage{titletoc}
\usepackage{mathrsfs}
\usepackage{upquote}
\usepackage[linewidth=0.5pt,leftmargin=1em,rightmargin=1em,innertopmargin=5pt,
    innerbottommargin=5pt,
    innerrightmargin=5pt,
    innerleftmargin=5pt]{mdframed}
\usepackage[toc,page,header]{appendix}

\usepackage{times}
\usepackage{latexsym}
\usepackage{tabularx}
\usepackage{multirow}
\usepackage{booktabs} 
\usepackage{enumitem}
\usepackage{amsmath}

\usepackage{inconsolata}
\usepackage{graphicx}

\usepackage[utf8]{inputenc} 
\usepackage[T1]{fontenc}    
\usepackage{hyperref}       
\usepackage{url}            
\usepackage{booktabs}       
\usepackage{amsfonts}       
\usepackage{nicefrac}       
\usepackage{microtype}      
\usepackage{xcolor}         
\usepackage{multicol}
\usepackage{transparent}
\usepackage{array}
\usepackage{bm}

\definecolor{nblue}{cmyk}{0.95,0.0,0.2,0.2}

\newcommand{\method}{\texttt{Hephaestus}\xspace}
\newcommand{\dataset}{\texttt{Hephaestus-Forge}\xspace}
\usepackage{xspace}
\usepackage{cleveref}
\usepackage{colortbl}
\usepackage{listings}
\definecolor{green}{HTML}{009B55}

\usepackage{listings}
\usepackage{pythonhighlight}
\usepackage{fancyvrb}
\RecustomVerbatimCommand{\VerbatimInput}{VerbatimInput}{fontsize=\footnotesize,
 frame=single,  
 framesep=0.5em, 
 labelposition=topline,
}

\usepackage{tcolorbox}
\tcbuselibrary{listings}

\crefname{section}{\S}{\S\S}
\Crefname{section}{\S}{\S\S}
\title{\method: Improving Fundamental Agent Capabilities of Large Language Models Through Continual Pre-Training}

\author{
\textbf{Yuchen Zhuang}$^1$\thanks{Work done during Yuchen's internship at Amazon. Correspondence to: Yuchen Zhuang (\texttt{yczhuang@gatech.edu}), Jingfeng Yang (\texttt{jingfengyangpku@gmail.com}), Chao Zhang (\texttt{chaozhang@gatech.edu}).} \quad \textbf{Jingfeng Yang}$^2$ \quad \textbf{Haoming Jiang}$^2$ \quad \textbf{Xin Liu}$^2$ \quad \textbf{Kewei Cheng}$^2$ \quad \\
\textbf{Sanket Lokegaonkar}$^2$ \quad  \textbf{Yifan Gao}$^2$ \quad \textbf{Qing Ping}$^2$ \quad \textbf{Tianyi Liu}$^2$ \quad \textbf{Binxuan Huang}$^2$ \quad \\ 
\textbf{Zheng Li}$^2$ \quad \textbf{Zhengyang Wang}$^2$ \quad \textbf{Pei Chen}$^2$ \quad \textbf{Ruijie Wang}$^2$ \quad \textbf{Rongzhi Zhang}$^1$ \quad \\ 
\textbf{Nasser Zalmout}$^2$ \quad \textbf{Priyanka Nigam}$^2$ \quad \textbf{Bing Yin}$^2$ \quad \textbf{Chao Zhang}$^1$ \\
$^1$ Georgia Institute of Technology  \quad  $^2$ Amazon \\
}

\begin{document}
\maketitle
\begin{abstract}

Due to the scarcity of agent-oriented pre-training data, LLM-based autonomous agents typically rely on complex prompting or extensive fine-tuning, which often fails to introduce new capabilities while preserving strong generalizability.
We introduce \dataset, the first large-scale pre-training corpus designed to enhance the fundamental capabilities of LLM agents in API function calling, intrinsic reasoning and planning, and adapting to environmental feedback. 
\dataset comprises 103B agent-specific data encompassing 76,537 APIs, including both tool documentation to introduce knowledge of API functions and function calling trajectories to strengthen intrinsic reasoning.
To explore effective training protocols, we investigate scaling laws to identify the optimal recipe in data mixing ratios. 
By continual pre-training on \dataset, \method outperforms small- to medium-scale open-source LLMs and rivals commercial LLMs on three agent benchmarks, demonstrating the effectiveness of our pre-training corpus in enhancing fundamental agentic capabilities and generalization of LLMs to new tasks or environments.



\end{abstract}

\section{Introduction}
\label{sec:intro}

\begin{figure}[t]
  \centering
  \includegraphics[width=\linewidth]{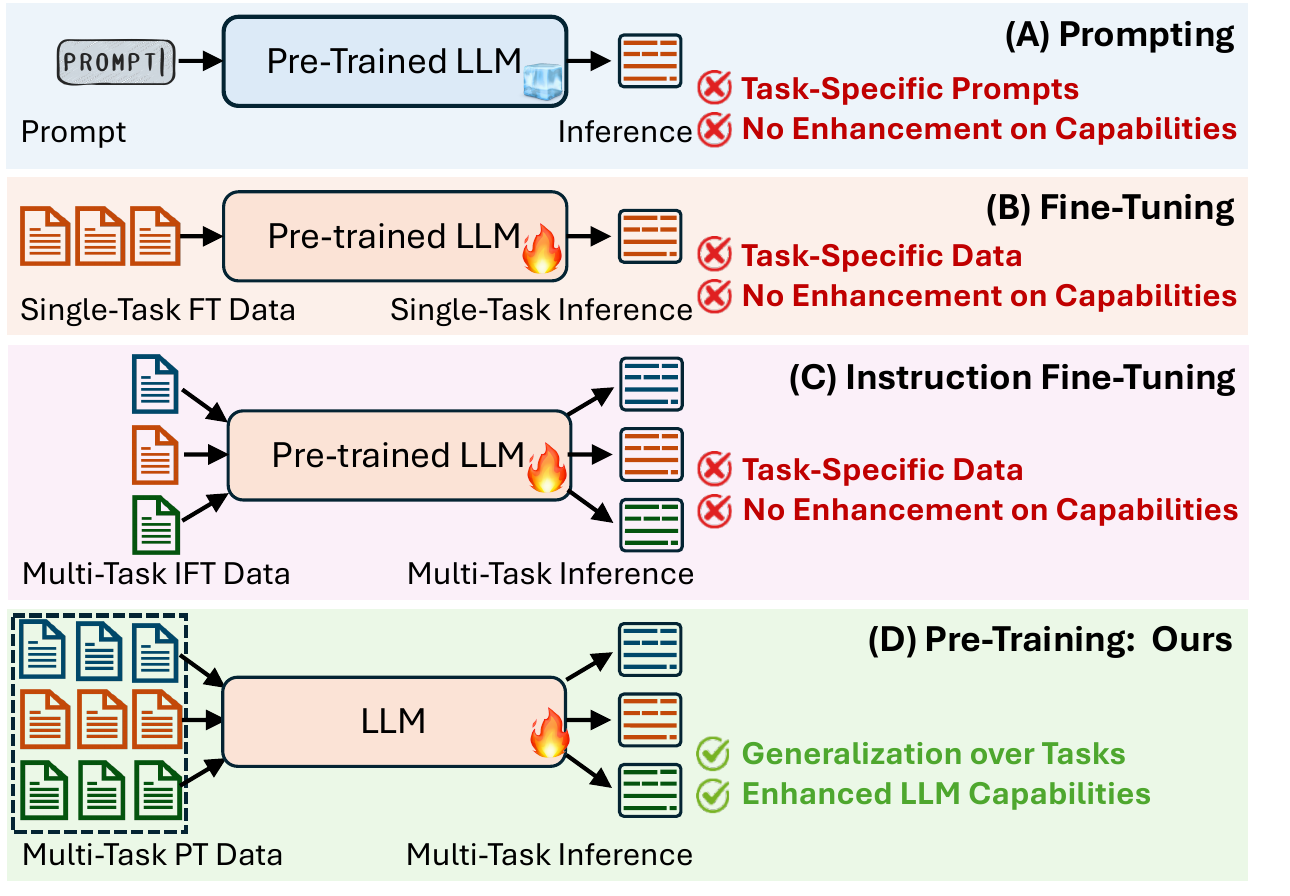}
  \caption{Training paradigms of LLM agents.
  \emph{Prompting} alone fails to introduce new knowledge and capabilities, while heavy \emph{fine-tuning} can hinder generalization and degrade performance in non-agent use cases, potentially suppressing the original base model capabilities.
  }
  \vspace{-2ex}
  \label{fig:related}
\end{figure} 

Large language models (LLMs) are rapidly evolving beyond traditional natural language processing tasks~\cite{ouyang2022training, brown2020language, achiam2023gpt}, demonstrating increasing intelligence and autonomy by exhibiting capabilities in perception, reasoning, planning, and action within complex real-world environments~\cite{yao2022react,lu2024chameleon,sun2024adaplanner}.
Through well-crafted prompting or extensive post-training, LLM-based autonomous agents augmented with external tools (\eg, APIs) have demonstrated exceptional instruction-following capabilities in a wide range of  tasks~\citep{schick2024toolformer,qin2023toolllm,srinivasan2023nexusraven,zeng2023agenttuning}. 

Despite their remarkable task-specific performance, existing LLM agents often face the following challenges: 
(1) \textbf{Overemphasis on instruction fine-tuning while ignoring the pre-training stage.} LLMs typically undergo a two-stage training process: pre-training to learn general knowledge and instruction fine-tuning to align to specific tasks and user preferences.
The \emph{Superficial Alignment Hypothesis}~\citep{zhou2024lima, gudibande2024the, lin2024the} posits that LLMs acquire most of their knowledge during pre-training, which is more important than instruction fine-tuning in terms of obtaining generalizable fundamental capabilities. 
However, the majority of existing agent frameworks (Figure~\ref{fig:related}) focus on instruction fine-tuning to align with specific patterns or formats, rather than fundamentally enhancing model knowledge or capabilities (\eg, API function calling).
(2) \textbf{Scarcity of agent-oriented pre-training data.} 
Agent instructions and trajectories significantly differ from general instructions and responses~\cite{zhang2024xlam}. Thus, function-calling knowledge is difficult to derive directly from web archives, the primary pre-training data source.
This notable lack of agent-specific pre-training corpora constrains LLMs from effectively acquiring new agentic knowledge and capabilities (Table~\ref{tab:related}).
(3) \textbf{Limited generalization across multiple tasks.}
LLM agents often struggle to generalize to new scenarios (\eg, from single to multiple tools) that differ from their original fine-tuning data distributions~\cite{qin2023toolllm}. 

To address these challenges, we introduce \dataset, a large-scale pre-training corpus specifically designed to enhance the fundamental capabilities of LLM agents in API function calling, intrinsic reasoning and planning, and adaptation to environmental feedback. 
Specifically, we focus on two primary objectives: (a) improving \emph{comprehension of individual function calls}, and (b) strengthening \emph{intrinsic reasoning capabilities} for solving problems requiring multiple function calls. 
To enhance (a) comprehension of API functions and alignment with their formats, we collect a large-scale dataset of tool documentation tailored for LLM pre-training on API function calls. 
Given the expanding range of tasks with growing complexity, we incorporate a vast number of function calling trajectories to improve (b) intrinsic reasoning abilities in sequencing API function calls. 
We then integrate this meticulously curated tool documentation and function-calling data with code (to bolster reasoning capabilities) and text data (to maintain robust text generation capabilities), creating a \emph{multi-source}, \emph{large-scale}, and \emph{high-quality} training corpus, \dataset.

Building upon \dataset, we introduce a continual pre-trained open-source LLM, \method, an LLM with strong agentic and autonomous capabilities across domains, bringing open-source models closer to the capabilities of commercial LLMs. 
Our empirical evaluations demonstrate that \texttt{\method-8B} outperforms open-source LLMs at small to medium scales (\eg, $9.6\%$ over \texttt{LLaMA-3-8B} and $17.6\%$ over \texttt{Mixtral-8x22B}) and performs comparably to API-based large commercial LLMs (\eg, $18.9\%$ over \texttt{Claude-3-Haiku} and $4.1\%$ over \texttt{GPT-3.5-turbo}) across three agent benchmarks. 
Our large-scale ablation studies further demonstrate the effectiveness of retrieved agent data in scaling up and diversifying the coverage of scenarios in pre-training.
Our contributions can be summarized as follows:
\begin{itemize}[leftmargin=0.6cm]
\setlength\itemsep{-0.2em}
\item We curate \dataset, a large-scale pre-training corpus designed to enhance understanding of API function calls and guide actionable trajectories for LLM agents. 
Remarkably, through exhaustive scaling law experiments, we discover a pioneering pre-training recipe with an empirically optimal data mix ratio. 
\item We propose \method, a foundation model that exhibits enhanced fundamental agentic capabilities, including API function calling, intrinsic reasoning and planning, and adaptation to environmental feedback, achieved through continual pre-training on \dataset.
\item We extensively compare \method with strong baselines across three agent benchmarks, verifying its enhanced fundamental agentic capabilities and superior generalization derived from \dataset. 
\end{itemize}


\section{Related Work}
\label{sec:relatedworks}

\begin{table*}[ht]
\begin{threeparttable}
\centering 
\renewcommand\arraystretch{0.98}
\fontsize{7}{9}\selectfont \setlength{\tabcolsep}{0.2em}
\begin{tabular}{@{}l|c|c|ccc|cc|cc|cccc@{}}
\toprule
\textbf{Methods} & \textbf{Datasets}           & \begin{tabular}[c]{@{}c@{}}\textbf{Training}\\ \textbf{Paradigm}\end{tabular} & \begin{tabular}[c]{@{}c@{}}\textbf{\# PT Data}\\ \textbf{(Tokens)}\end{tabular} & \begin{tabular}[c]{@{}c@{}}\textbf{\# IFT Data}\\ \textbf{(Samples)}\end{tabular} & \textbf{\# APIs} & \textbf{Code}  & \begin{tabular}[c]{@{}c@{}}\textbf{Nat.}\\ \textbf{Lang.}\end{tabular} & \begin{tabular}[c]{@{}c@{}}\textbf{Action}\\ \textbf{Traj.}\end{tabular} & \begin{tabular}[c]{@{}c@{}}\textbf{API}\\ \textbf{Doc.}\end{tabular} & \begin{tabular}[c]{@{}c@{}}\textbf{Func.}\\ \textbf{Call}\end{tabular} & \begin{tabular}[c]{@{}c@{}}\textbf{Multi.}\\ \textbf{Step}\end{tabular}  & \begin{tabular}[c]{@{}c@{}}\textbf{Plan}\\ \textbf{Refine}\end{tabular}  & \begin{tabular}[c]{@{}c@{}}\textbf{Multi.}\\ \textbf{Turn}\end{tabular}\\ \midrule 
\multicolumn{13}{l}{\emph{Instruction Finetuning-based LLM Agents for Intrinsic Reasoning}}  \\ \midrule
FireAct~\cite{chen2023fireact} & FireAct & IFT & - & 2.1K & 10 & \textcolor{red}{\XSolidBrush} &\textcolor{green}{\CheckmarkBold} &\textcolor{green}{\CheckmarkBold}  & \textcolor{red}{\XSolidBrush} &\textcolor{green}{\CheckmarkBold} & \textcolor{red}{\XSolidBrush} &\textcolor{green}{\CheckmarkBold} & \textcolor{red}{\XSolidBrush} \\
ToolAlpaca~\cite{tang2023toolalpaca} & ToolAlpaca & IFT & - & 4.0K & 400 & \textcolor{red}{\XSolidBrush} &\textcolor{green}{\CheckmarkBold} &\textcolor{green}{\CheckmarkBold} & \textcolor{red}{\XSolidBrush} &\textcolor{green}{\CheckmarkBold} & \textcolor{red}{\XSolidBrush}  &\textcolor{green}{\CheckmarkBold} & \textcolor{red}{\XSolidBrush}  \\
ToolLLaMA~\cite{qin2023toolllm} & ToolBench & IFT & - & 12.7K & 16,464 & \textcolor{red}{\XSolidBrush} &\textcolor{green}{\CheckmarkBold} &\textcolor{green}{\CheckmarkBold} &\textcolor{red}{\XSolidBrush} &\textcolor{green}{\CheckmarkBold}&\textcolor{green}{\CheckmarkBold}&\textcolor{green}{\CheckmarkBold} &\textcolor{green}{\CheckmarkBold}\\
AgentEvol~\citep{xi2024agentgym} & AgentTraj-L & IFT & - & 14.5K & 24 &\textcolor{red}{\XSolidBrush} & \textcolor{green}{\CheckmarkBold} &\textcolor{green}{\CheckmarkBold}&\textcolor{red}{\XSolidBrush} &\textcolor{green}{\CheckmarkBold}&\textcolor{red}{\XSolidBrush} &\textcolor{red}{\XSolidBrush} &\textcolor{green}{\CheckmarkBold}\\
Lumos~\cite{yin2024agent} & Lumos & IFT  & - & 20.0K & 16 &\textcolor{red}{\XSolidBrush} & \textcolor{green}{\CheckmarkBold} & \textcolor{green}{\CheckmarkBold} &\textcolor{red}{\XSolidBrush} & \textcolor{green}{\CheckmarkBold} & \textcolor{green}{\CheckmarkBold} &\textcolor{red}{\XSolidBrush} & \textcolor{green}{\CheckmarkBold}\\
Agent-FLAN~\cite{chen2024agent} & Agent-FLAN & IFT & - & 24.7K & 20 &\textcolor{red}{\XSolidBrush} & \textcolor{green}{\CheckmarkBold} & \textcolor{green}{\CheckmarkBold} &\textcolor{red}{\XSolidBrush} & \textcolor{green}{\CheckmarkBold}& \textcolor{green}{\CheckmarkBold}&\textcolor{red}{\XSolidBrush} & \textcolor{green}{\CheckmarkBold}\\
AgentTuning~\citep{zeng2023agenttuning} & AgentInstruct & IFT & - & 35.0K & - &\textcolor{red}{\XSolidBrush} & \textcolor{green}{\CheckmarkBold} & \textcolor{green}{\CheckmarkBold} &\textcolor{red}{\XSolidBrush} & \textcolor{green}{\CheckmarkBold} &\textcolor{red}{\XSolidBrush} &\textcolor{red}{\XSolidBrush} & \textcolor{green}{\CheckmarkBold}\\\midrule
\multicolumn{13}{l}{\emph{Instruction Finetuning-based LLM Agents for Function Calling}} \\\midrule
NexusRaven~\citep{srinivasan2023nexusraven} & NexusRaven & IFT & - & - & 116 & \textcolor{green}{\CheckmarkBold} & \textcolor{green}{\CheckmarkBold}  & \textcolor{green}{\CheckmarkBold} &\textcolor{red}{\XSolidBrush} & \textcolor{green}{\CheckmarkBold} &\textcolor{red}{\XSolidBrush} &\textcolor{red}{\XSolidBrush}&\textcolor{red}{\XSolidBrush}\\
Gorilla~\citep{patil2023gorilla} & Gorilla & IFT & - & 16.0K & 1,645 & \textcolor{green}{\CheckmarkBold} &\textcolor{red}{\XSolidBrush} &\textcolor{red}{\XSolidBrush}&\textcolor{green}{\CheckmarkBold} &\textcolor{green}{\CheckmarkBold} &\textcolor{red}{\XSolidBrush} &\textcolor{red}{\XSolidBrush} &\textcolor{red}{\XSolidBrush}\\
OpenFunctions-v2~\citep{patil2023gorilla} & OpenFunctions-v2 & IFT & - & 65.0K & - & \textcolor{green}{\CheckmarkBold} & \textcolor{green}{\CheckmarkBold} &\textcolor{red}{\XSolidBrush} &\textcolor{green}{\CheckmarkBold} &\textcolor{green}{\CheckmarkBold} &\textcolor{red}{\XSolidBrush} &\textcolor{red}{\XSolidBrush} &\textcolor{red}{\XSolidBrush}\\
API Pack~\cite{guo2024api} & API Pack & IFT & - & 1.1M & 11,213 &\textcolor{green}{\CheckmarkBold} &\textcolor{red}{\XSolidBrush} &\textcolor{green}{\CheckmarkBold} &\textcolor{red}{\XSolidBrush} &\textcolor{green}{\CheckmarkBold} &\textcolor{red}{\XSolidBrush}&\textcolor{red}{\XSolidBrush}&\textcolor{red}{\XSolidBrush}\\ 
LAM~\citep{zhang2024agentohana} & AgentOhana & IFT & - & 42.6K & - & \textcolor{green}{\CheckmarkBold} & \textcolor{green}{\CheckmarkBold} &\textcolor{green}{\CheckmarkBold}&\textcolor{red}{\XSolidBrush} &\textcolor{green}{\CheckmarkBold}&\textcolor{red}{\XSolidBrush}&\textcolor{green}{\CheckmarkBold}&\textcolor{green}{\CheckmarkBold}\\
xLAM~\citep{liu2024apigen} & APIGen & IFT & - & 60.0K & 3,673 & \textcolor{green}{\CheckmarkBold} & \textcolor{green}{\CheckmarkBold} &\textcolor{green}{\CheckmarkBold}&\textcolor{red}{\XSolidBrush} &\textcolor{green}{\CheckmarkBold}&\textcolor{red}{\XSolidBrush}&\textcolor{green}{\CheckmarkBold}&\textcolor{green}{\CheckmarkBold}\\\midrule
\multicolumn{13}{l}{\emph{Pretraining-based LLM Agents}}  \\\midrule
\rowcolor{teal!12} \method & \dataset & PT & 103B & 95.0K  & 76,537  & \textcolor{green}{\CheckmarkBold} & \textcolor{green}{\CheckmarkBold} & \textcolor{green}{\CheckmarkBold} & \textcolor{green}{\CheckmarkBold} & \textcolor{green}{\CheckmarkBold} & \textcolor{green}{\CheckmarkBold} & \textcolor{green}{\CheckmarkBold} & \textcolor{green}{\CheckmarkBold}\\
\bottomrule
\end{tabular}
\caption{Summary of existing instruction finetuning-based LLM agents for intrinsic reasoning and function calling, along with their training resources and sample sizes. "PT" and "IFT" denote "Pre-Training" and "Instruction Fine-Tuning", respectively.}
\vspace{-2ex}
\label{tab:related}
\end{threeparttable}
\end{table*}

\noindent \textbf{Prompting-based LLM Agents.} Due to the lack of agent-specific pre-training corpus, existing LLM agents rely on either prompt engineering~\cite{hsieh2023tool,lu2024chameleon,yao2022react,wang2023voyager} or instruction fine-tuning~\cite{chen2023fireact,zeng2023agenttuning} to understand human instructions, decompose high-level tasks, generate grounded plans, and execute multi-step actions. 
However, prompting-based methods mainly depend on the capabilities of backbone LLMs (usually commercial LLMs), failing to introduce new knowledge and struggling to generalize to unseen tasks~\cite{sun2024adaplanner,zhuang2023toolchain}. 

\noindent \textbf{Instruction Finetuning-based LLM Agents.} Considering the extensive diversity of APIs and the complexity of multi-tool instructions, tool learning inherently presents greater challenges than natural language tasks, such as text generation~\cite{qin2023toolllm}.
Post-training techniques focus more on instruction following and aligning output with specific formats~\cite{patil2023gorilla,hao2024toolkengpt,qin2023toolllm,schick2024toolformer}, rather than fundamentally improving model knowledge or capabilities. 
Moreover, heavy fine-tuning can hinder generalization or even degrade performance in non-agent use cases, potentially suppressing the original base model capabilities~\cite{ghosh2024a}.

\noindent \textbf{Pretraining-based LLM Agents.} While pre-training serves as an essential alternative, prior works~\cite{nijkamp2023codegen,roziere2023code,xu2024lemur,patil2023gorilla} have primarily focused on improving task-specific capabilities (\eg, code generation) instead of general-domain LLM agents, due to single-source, uni-type, small-scale, and poor-quality pre-training data. 
Existing tool documentation data for agent training either lacks diverse real-world APIs~\cite{patil2023gorilla, tang2023toolalpaca} or is constrained to single-tool or single-round tool execution. 
Furthermore, trajectory data mostly imitate expert behavior or follow function-calling rules with inferior planning and reasoning, failing to fully elicit LLMs' capabilities and handle complex instructions~\cite{qin2023toolllm}. 
Given a wide range of candidate API functions, each comprising various function names and parameters available at every planning step, identifying globally optimal solutions and generalizing across tasks remains highly challenging.

\section{Preliminaries}
\label{sec:preliminary}

\begin{figure*} [t]
	\centering
   \vspace{-2ex}
	\subfigure[\dataset]{
		\includegraphics[width=0.225\linewidth]{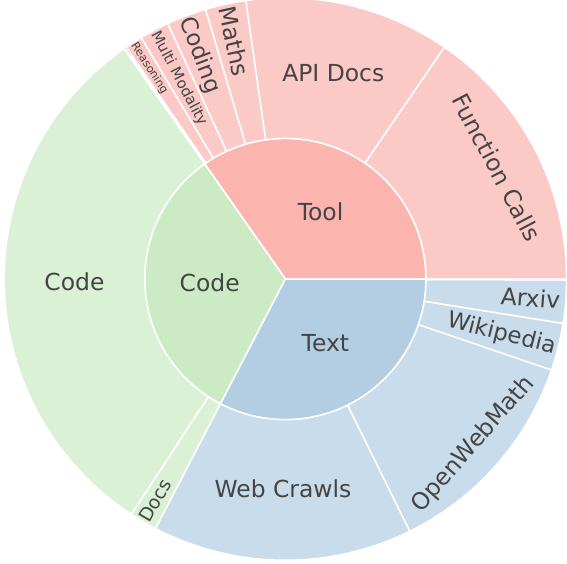}
		\label{fig:data-overall}
	} 
     \subfigure[Tool Data]{
		\includegraphics[width=0.225\linewidth]{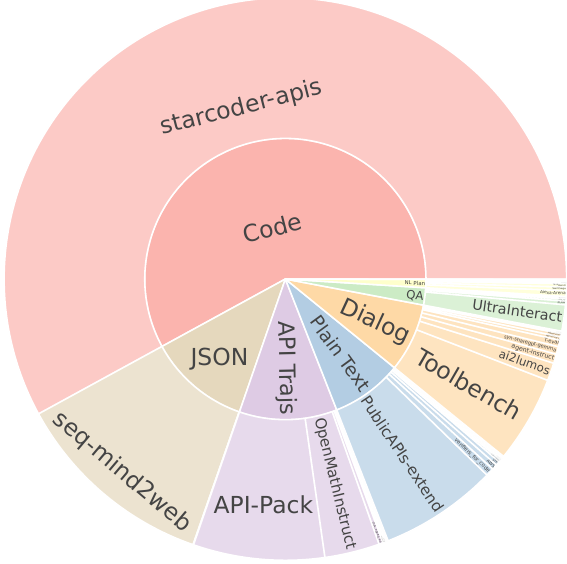}
		\label{fig:data-tool}
	}
    \subfigure[Retrieved Data]{
		\includegraphics[width=0.225\linewidth]{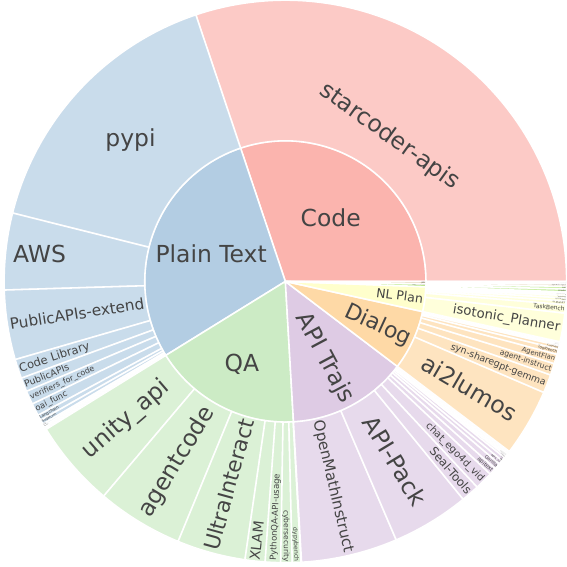}
		\label{fig:data-retr}
	}
    \subfigure[t-SNE: Retrieved Data]{
		\includegraphics[width=0.225\linewidth]{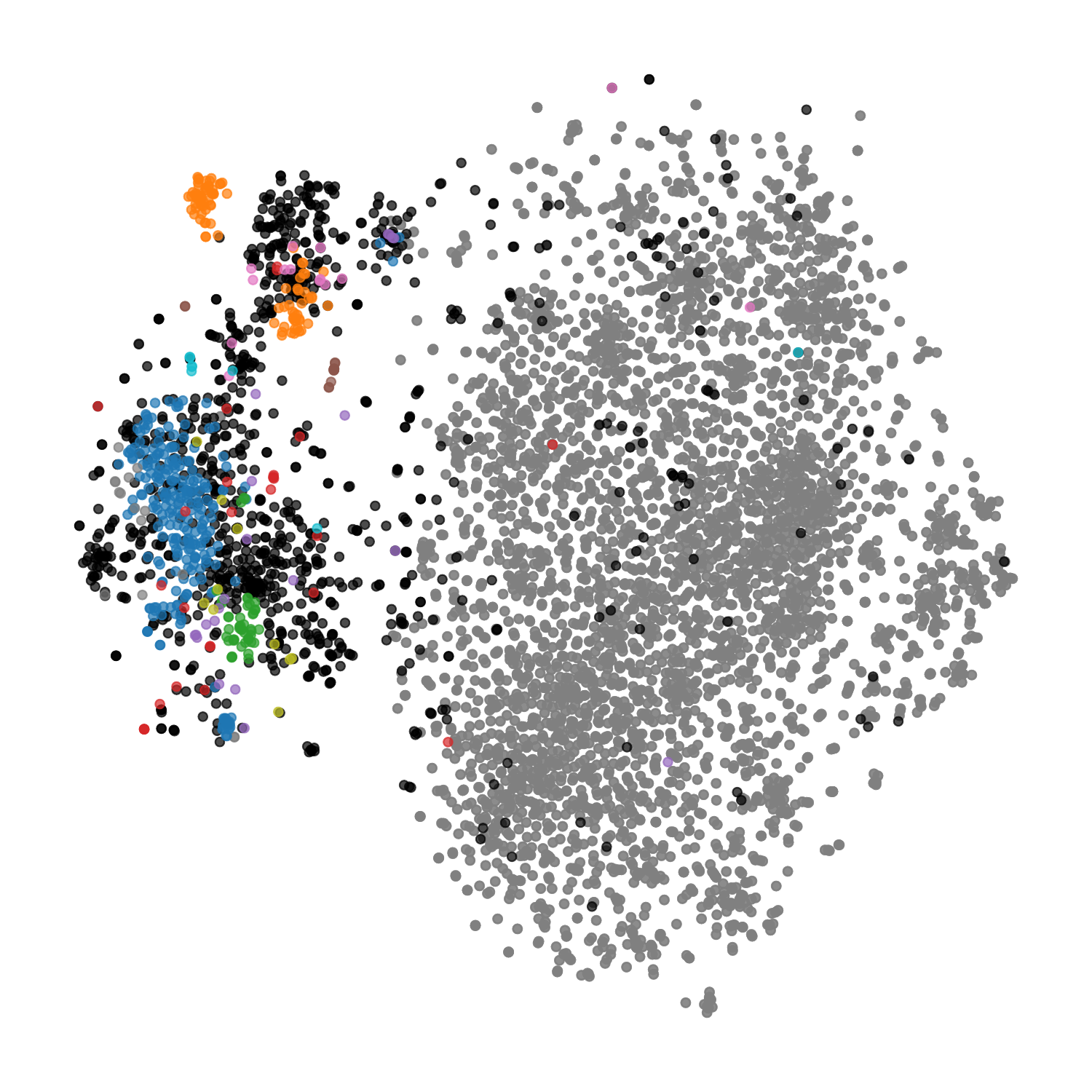}
		\label{fig:tsne-retr}
	}
	\caption{Data composition of (a) the entire \dataset, (b) seed data collection (\cref{sec:data-phase1}), and (c) retrieved agent data from the open web (\cref{sec:data-phase2}). A t-SNE visualization (d) depicts seed data (\textbf{colorful} points, with each color representing different data sources), retrieved data (\textbf{black}), and general text (\textcolor{gray}{\textbf{gray}}) within the semantic space, where retrieved data is closer to the selected seed data than to the general text. Detailed data sources are in \cref{app:data-pretrain}.
 }
\vspace{-2ex}
\label{fig:data}
\end{figure*}

\noindent \textbf{Problem Formulation.} We conceptualize leveraging LLMs as autonomous agents for problem-solving as a planning process.
Initially, we augment the LLM agent with access to a pool of candidate API functions, denoted as $\mathcal{A}=\{\text{API}_0,\text{API}_1,\cdots,\text{API}_m$\}, along with a natural language task description $g\in\mathcal{G}$ from the task space $\mathcal{G}$. 
The objective of the LLM agent is to translate the task description $g$ into an ordered sequence of $T_g$ API function calls $p_g=\{a_0,\cdots,a_{T_g}\}$.
Specifically, considering the task description $g$ as the initial state $s_0$, we then sample the plan $p_g$ by prompting the LLM agent with the API definitions $\mathcal{I}$ and demonstration samples $\mathcal{D}$ as follows: $p_g\sim\rho(a_0,a_1,\cdots,a_{T_g}|s_0;\mathcal{I},\mathcal{D}):\mathcal{G}\times\mathcal{I}\times\mathcal{D}\to\Delta(\mathcal{A}^{T_g})$, where $\Delta(\cdot)$ denotes a probability simplex function. 
The final output is derived after executing the entire plan $y\sim\pi(y|s_0,a_1,a_2,\cdots,a_{T_g})$, where $\pi(\cdot)$ denotes a plan executor.

During this procedure, we focus on three fundamental capabilities of LLM agents:

\noindent \textbf{Accurate Function Calling.} It involves accurately understanding the API definitions and demonstration samples to generate correct API function calls with corresponding parameters in a given scenario.
Specifically, the model should accurately understand the API definitions $\mathcal{I}$ and demonstration samples $\mathcal{D}$, as well as generate an accurate API function call in the given scenario $p(a_t|s_0,a_1,\cdots,a_{t-1},\mathcal{I},\mathcal{D})$, where $a_t$ is the ground-truth API function call with corresponding parameters at $t$-th step.

\noindent \textbf{Intrinsic Reasoning and Planning.} It refers to the intrinsic reasoning and planning ability to devise a sequence of multiple tool functions as a solution when addressing complex (multi-step) real-world problems. In such cases, LLMs are often required to generate a sequence of API function calls, $p(a_1,a_2,\cdots,a_{T_g}|s_0;\mathcal{I},\mathcal{D})$, where $\{a_1,a_2,\cdots,a_{T_g}\}$ constitutes the ground-truth solution plan of length $T_g$.  
This process relies on intrinsic reasoning embedded within the model parameters; enhanced reasoning capabilities lead to a solution plan with a higher chance of success.

\noindent \textbf{Adaptation with Environment Feedback.} It focuses on adapting the current plan or action based on environmental feedback when the environments support interaction with the LLM agent. When such feedback is available, it is crucial for the agent to adjust its actions accordingly: $p(a_t|s_0,a_1,o_1,a_2,\cdots,o_{t-1};\mathcal{I},\mathcal{D})$,
where $o_{k}$ represents the feedback from the environment after the $k$-th action. 
Incorporating environmental feedback allows the agent to take reflections to refine its plan and improve task performance iteratively.

\section{\dataset}
\label{sec:dataset}

To scale and diversify the pre-training corpus for LLM agents, we introduce a three-stage construction process for \dataset (see Figure~\ref{fig:data}): (1) \textbf{Seed Data Collection} (\cref{sec:data-phase1}), where we gather initial high-quality samples; (2) \textbf{Web Data Retrieval} (\cref{sec:data-phase2}), which expands the seed data by retrieving relevant data from the web; and (3) \textbf{Data Quality Control} (\cref{sec:data-phase3}), where we ensure the integrity and relevance of the collected data. 

\subsection{Seed Data Collection}
\label{sec:data-phase1}

For seed data collection, we first traverse available public resources to gather high-quality API documentation and action trajectories, including: 
(1) \textbf{Public APIs.} High-quality API documentation is collected from over $1,400$ public APIs and official websites, including detailed function definitions and parameter descriptions. 
(2) \textbf{Public Repositories.} To improve intrinsic reasoning, we integrate action trajectories from over $60$ public repositories across diverse domains, such as programming code and web interactions. 
(3) \textbf{Code-to-Text Synthesis.} Given the limited coverage of curated data, we use LLMs to synthesize additional API documentation from \emph{StarCoder-API}, generating examples based on code snippets. 
(4) \textbf{Simulated Agent Data.} We gather simulated action sequences with observational data to facilitate adaptation to environmental feedback. 
Importantly, we offer step-by-step details of the seed data collection process in \cref{app:4-1} for reproducibility.

\subsection{Web Data Retrieval}
\label{sec:data-phase2}
Given the limited availability of agent-oriented data, we use the high-quality data described in \cref{sec:data-phase1} as seed data for further expansion. To enhance agentic capabilities, we retrieve a diverse set of examples from web crawls, focusing on content relevant to API documentation and action trajectories.
Our retrieval process involves the following steps:
(1) \textbf{Web Data Corpus Creation.} Similar to CommonCrawl~\cite{raffel2020exploring} and FineWeb~\citep{penedo2024finewebdatasetsdecantingweb}, we first compile a large-scale web data corpus.
(2) \textbf{Semantic Matching.} We utilize COCO-DR~\cite{yu2022coco} to encode semantic representations of documents in the seed data and the large-scale web corpus. 
We then retrieve the top-$K$ similar documents by calculating the cosine similarity between the corresponding embeddings. 
It allows us to identify and retrieve documents from the web corpus that are semantically similar to our seed data, effectively enriching our dataset with relevant and diverse information.
(3) \textbf{Quality Control.} To ensure the quality of the retrieved corpus, we perform data pruning to remove semantically redundant content and maintain the diversity of knowledge, preventing overrepresentation of certain topics and ensuring generalization and robustness across domains.

\subsection{Data Quality}
\label{sec:data-phase3}
After retrieving semantically relevant data from the web corpus, we obtain a collection of noisy agent data. To ensure the integrity and relevance of our dataset, it is essential to consistently monitor data quality and filter out content that resembles general text rather than agent-specific data.
First, we employ \texttt{Claude-3-Sonnet}~\cite{claude-3} as the data annotator to annotate a total of $71,473$ samples from the retrieved data, identifying $37,714$ as agent-relevant and $33,767$ as general text paragraphs.
Using the annotated samples, we train a \texttt{fastText}~\cite{joulin2016fasttext} model to effectively recall additional agent-relevant web data. 
This filtering process then reduces the data volume from approximately $200$B to $80$B tokens, ensuring that the preserved data maintains high relevance and quality. See details in \cref{app:4-3}.

\begin{figure}[t]
  \centering
  \includegraphics[width=\linewidth]{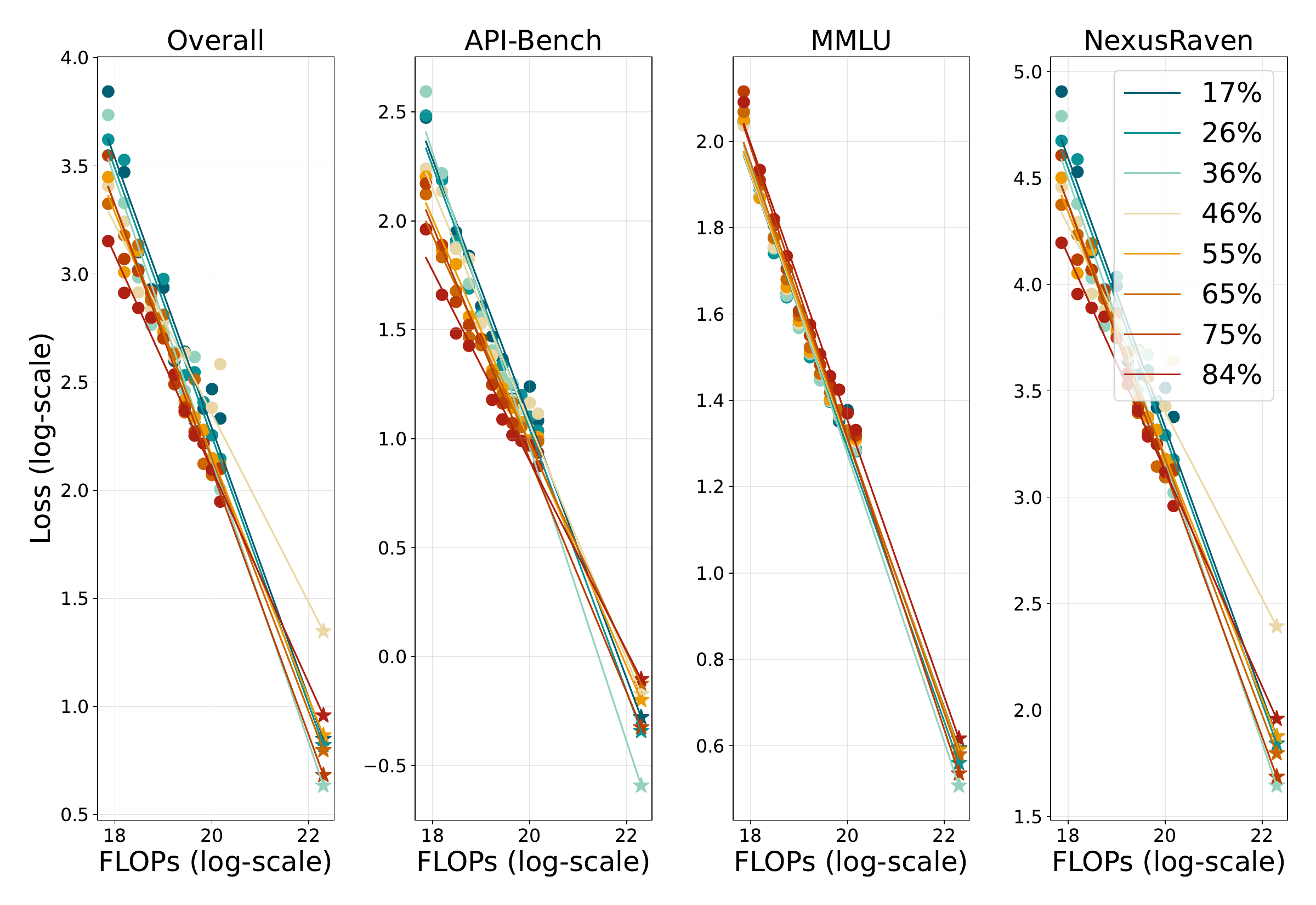}
  \caption{Scaling law of the relationship between agent data mixing ratio ($\%$) and benchmark loss. 
  }
  \label{fig:sl}
\end{figure}

\begin{figure*}[t]
  \centering
  \includegraphics[width=\linewidth]{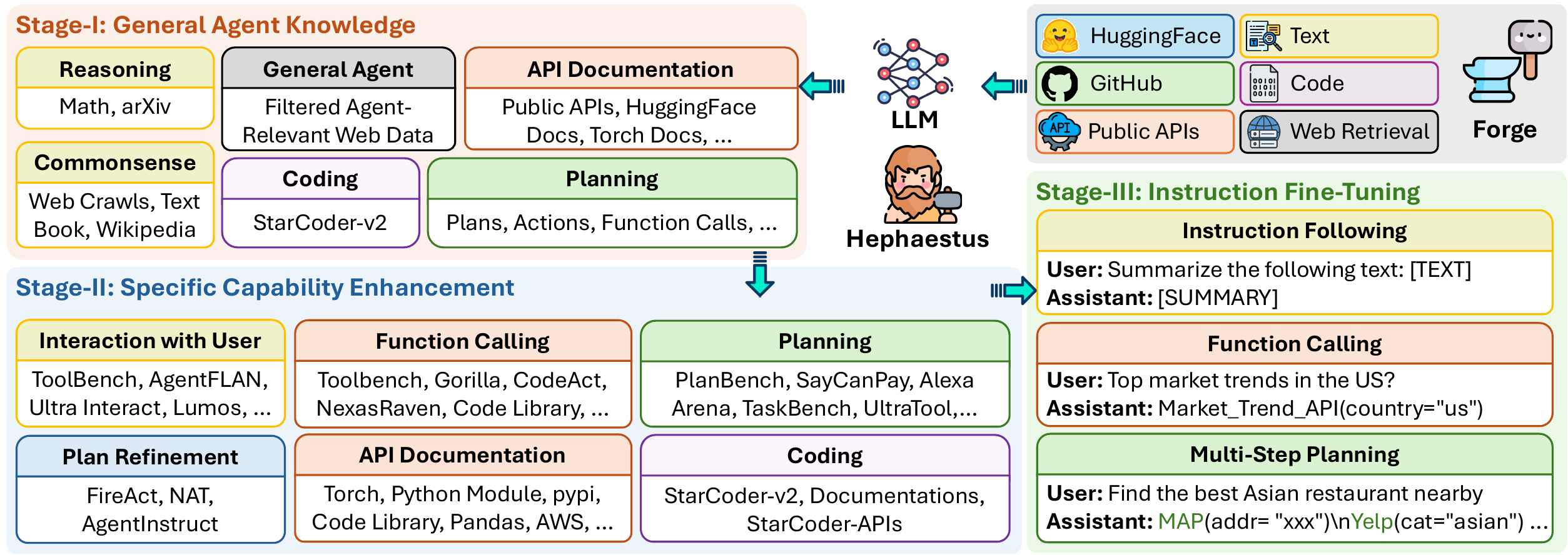}
  \caption{
  Overview of the pre-training (Stages I \& II) and instruction fine-tuning (III) framework in \method.
  }
  \vspace{-2ex}
  \label{fig:overview}
\end{figure*}

\section{Scaling Laws for Data Composition}
\label{sec:data-phase4}

When designing LLMs, the scaling law~\cite{kaplan2020scaling, hoffmann2022training} is an important predictive tool that can estimate the performance (\eg, benchmark loss) of a large-sized target model using a scaling curve fitted over much smaller models (referred to as sampling models).
We develop scaling laws to determine the optimal data proportion among agent data, text data, and code data. With the total budget of the data volume fixed, our scaling law experiments show that the effect of agent data ratio $x$ on the loss $\mathcal{L}$ of a pre-trained model follows power laws:
\begin{equation*}
    \begin{aligned}
        \mathcal{L}=c+kx^{\alpha},
    \end{aligned}
\end{equation*}
where $c$, $k$, and $\alpha$ are parameters to be fitted.
By fitting these parameters using a collection of small models, training data, or computational resources, scaling laws can extrapolate to precisely predict the test loss of larger cases over orders of magnitude.

\noindent\textbf{Scaling Law Experiments.}
Concretely, we construct our scaling laws by pre-training models ranging in $45$M to $0.65$B parameters.
To simulate the continual pre-training setting, we amplify the target data volume used for training each small model to $50\times$ model parameters.
Consequently, the total compute budgets for the scaling law experiments span from $7\times10^{17}$ to $2\times10^{20}$ FLOPs.
Regarding data proportions, we begin with the seed agent data and progressively incorporate the retrieved web corpus to increase the agent data ratio. Concurrently, as the agent data ratio increases, we proportionally decrease the volumes of general text and code data to maintain the fixed total data volume.
Following~\citet{dubey2024llama}, we leverage the benchmark loss of Nexus~\cite{srinivasan2023nexusraven}, API-Bank~\cite{li2023api}, API-Bench~\cite{patil2023gorilla} to monitor the agent capabilities, and MMLU~\cite{hendrycks2020measuring} to monitor the general capabilities of LLMs.

\noindent\textbf{Optimal Data Mixing Ratio.}
Figure~\ref{fig:sl} illustrates that the optimal mixture of agent data within the entire pre-training corpus is approximately 36\%, indicating that the proportion of agent data, text data, and code data should be roughly $1:1:1$.
This balanced distribution promotes both specialized agent capabilities and general language understanding, ensuring that the model remains versatile and robust across diverse tasks and domains.

\noindent\textbf{Remark.} The established scaling laws provide critical insights into the data composition for pre-training LLM agents. By identifying the optimal ratio of agent data, we ensure that the model effectively balances specialized agentic capabilities with general language proficiency. 


\section{\method}
\label{sec:method}

In this section, we propose \method, a foundation model with enhanced fundamental capabilities of LLM agents.
\method undergoes a two-stage continual pre-training process, followed by instruction fine-tuning (see Figure~\ref{fig:overview}):
(1) \textbf{Stage I}, continual pre-training stage on the entire \dataset corpus to inject general agent knowledge (\cref{sec:method-pre});
(2) \textbf{Stage II}, continual pre-training stage on the high-quality seed set of \dataset to further enhance specific capabilities (\cref{sec:method-pre}); and
(3) \textbf{Stage III}, instruction fine-tuning to follow general instructions and downstream task requirements (\cref{sec:method-ift}).

\subsection{Stage I \& II: Continual Pre-Training}
\label{sec:method-pre}
Following~\citet{caccia2022new, lange2023continual}, we revisit the concept of \emph{stability gap}, which describes the phenomenon where the performance on old tasks initially drops and then recovers when learning a new task.
Specifically, in the continual pre-training of LLMs, if the data distribution shifts too significantly between the initial pre-training and the continual pre-training stages, the model's capabilities can deteriorate markedly until it assimilates knowledge from the new data distribution~\cite{guo2024efficient}.
To this end, we propose a two-stage continual pre-training framework:

\noindent \textbf{Stage I: Injecting General Agent Knowledge.}  
Stage I infuses general agent knowledge, accompanied by commonsense knowledge and code snippets. We pre-train \method on the entire \dataset, whose data distribution is carefully balanced between general corpus and agent-specific data, facilitating a smooth and gradual integration of agent knowledge.

\noindent \textbf{Stage II: Enhancing Agent-Specific Capabilities.}
Stage II leverages high-quality agent data to further enhance the specific capabilities of an agent LLM, including user interaction, function calling, planning, plan refinement, and coding capabilities. We continually pre-train the model obtained from Stage I on the high-quality seed data in \cref{sec:data-phase1} to further align the behavior with agent-specific requirements, ensuring that the specialized functionalities are robustly learned and integrated.

\noindent \textbf{Pre-Training Objectives.} For both stages, we employ language modeling as the primary pre-training task. 
The objective is to auto-regressively predict the next token, defined as follows:
\begin{equation*}
    \begin{aligned}
        \mathcal{L}_{\text{PT}}=-\mathbb{E}_{\mathbf{x}\in\mathcal{D}_{\text{PT}}}\sum_{i=1}^np(x_i|\mathbf{x}_{<i}),
    \end{aligned}
\end{equation*}
where $\mathcal{D}_{PT}$ denotes the pre-training data, and $x_i$ represents the $i$-th token in the training sample $\mathbf{x}$.

\subsection{Stage III: Instruction Fine-Tuning}
\label{sec:method-ift}
To further improve its instruction-following capabilities to align with complex agent environments, \method undergoes instruction fine-tuning on a blend of high-quality instruction-completion datasets, including \emph{ShareGPT}~\cite{chiang2023vicuna}, \emph{ToolACE}~\cite{liu2024toolace}, and \emph{AgentFlan}~\cite{chen2024agent}. 
The Stage III employs a negative log-likelihood loss function, defined as:
\begin{equation*}
    \begin{aligned}
        \mathcal{L}_{\text{IFT}}=-\mathbb{E}_{(\mathbf{x},\mathbf{y})\in\mathcal{D}_{\text{IFT}}}\sum_{i=1}^np(y_i|\mathbf{y}_{<i},\mathbf{x}),
    \end{aligned}
\end{equation*}
where $\mathbf{x}$ represents the given instruction, and $\mathbf{y}$ is the expected solution to fill. Here, $(\mathbf{x},\mathbf{y})\in\mathcal{D}_{\text{IFT}}$ indicates that the data pairs are sampled from the instruction-tuning dataset. 
\section{Experiments}
\label{sec:exp}

\begin{figure*}[t]
  \centering
  \vspace{-2ex}
  \includegraphics[width=\linewidth]{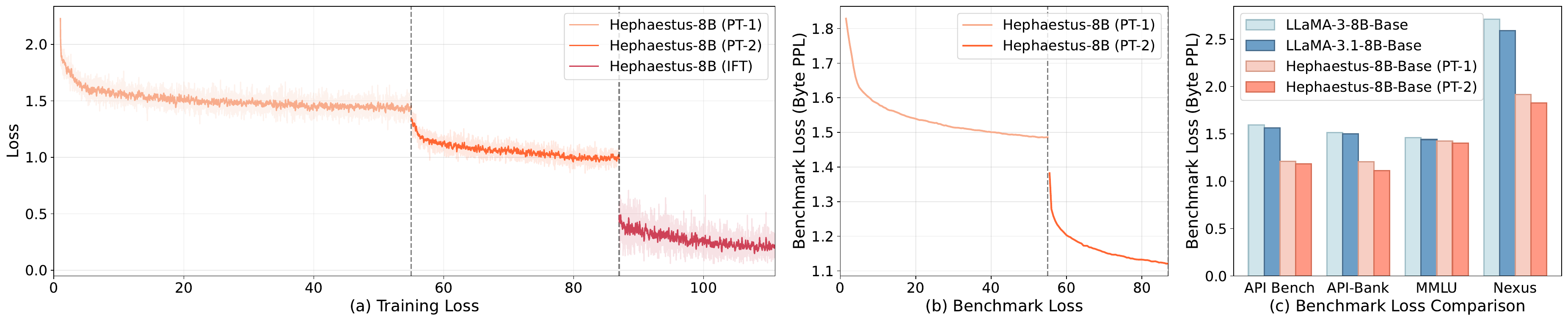}
  \caption{Training and benchmark loss. (a) Training loss of \method during continual pre-training and instruction fine-tuning. (b) Benchmark loss at periodic training checkpoints and (c) a comparison across base models.
  }
  \vspace{-1ex}
  \label{fig:val_loss}
\end{figure*}


\subsection{Experiment Setup}
\noindent \textbf{Tasks and Datasets.} 
We mainly evaluate our \method on the following benchmarks:
(1) \emph{AgentBench}~\cite{liu2024agentbench} for intrinsic reasoning and adaptation to environment feedback;
(2) \emph{Berkeley Function Calling Leaderboard (BFCL)-v3} and (3) \emph{BFCL-v2}~\cite{patil2023gorilla} for accurate function calling.
To test generalizability instead of memorization, we intentionally exclude all evaluation benchmarks from pre-training corpora.
Task and dataset details are available in \cref{app:data}.

\noindent \textbf{Baselines.}
We mainly compare to the following baselines: (1) \emph{Base LLMs} and 
(2) \emph{Open-Source Instruction Fine-Tuned LLMs} with varying model sizes. 
We also show the performance of (3) \emph{API-based Commercial LLMs} as reference. 
We exclude prompting and instruction fine-tuned agent frameworks from our main experiments to focus on evaluating the fundamental agentic capabilities of LLMs.
Details of baseline models are in \cref{app:baseline}.

\noindent \textbf{Evaluation.}
Following~\citet{liu2024agentbench,patil2023gorilla},
for AgentBench, we report \emph{success rate} for the OS, DB, HH, and WB environments, \emph{F1 score} for the KG environment, and \emph{reward score} for the WS environment; for BFCL-v2 and -v3, we use \emph{accuracy} as the primary metric for all scenarios to assess correct function calls.
Implementation details can be found in \cref{app:implementation}.


\begin{table*}[t]
\centering
\fontsize{7}{9}\selectfont\setlength{\tabcolsep}{0.4em}
\begin{tabular}{@{}lcc>{}ccccccc>{}ccccc>{}c@{}}
\toprule
\textbf{Datasets ($\rightarrow$)} & \multicolumn{2}{c}{\textbf{Model}} & \multicolumn{7}{c}{\textbf{AgentBench}} & \multicolumn{5}{c}{\textbf{BFCL-v3}}& \textbf{BFCL-v2}\\
\cmidrule(lr){2-3} \cmidrule(lr){4-10} \cmidrule(lr){11-15} \cmidrule(lr){16-16}
\textbf{Models ($\downarrow$)} & \textbf{Size} & \textbf{Type} & \textbf{OA} & \textbf{OS} & \textbf{DB} &  \textbf{HH} & \textbf{KG} & \textbf{WB} & \textbf{WS} & \textbf{OA} & \textbf{NL-AST} & \textbf{Exec} & \textbf{L-AST} & \textbf{MT} & \textbf{OA}\\\midrule
\multicolumn{16}{l}{\emph{Base LLMs}}  \\ \midrule
LLaMA-3-8B~\cite{dubey2024llama} & 8B & OSS & 0.56 & 2.8 & 12.0 & 0.0 & 8.9 & 11.0 & 1.4 & 17.73 & 4.3 & 2.5 & 39.1 & 0.0 & 17.77\\
LLaMA-3.1-8B~\cite{dubey2024llama} & 8B & OSS & 1.05 & 15.3 & 5.3 & 8.0 & 12.7 & \textbf{18.0} & 41.9 & 19.50 & 16.3 & 10.7 & 37.5 & 0.0 & 21.08 \\
\rowcolor{teal!12} \method-8B-Base & 8B & OSS & \textbf{1.87} & \textbf{20.8} & \textbf{32.3} & \textbf{30.0} & \textbf{16.0} & 16.0 & \textbf{60.5} & \textbf{22.12} & \textbf{18.1} & \textbf{12.1} & \textbf{42.2} & \textbf{4.0} & \textbf{25.18}\\\midrule
\multicolumn{16}{l}{\emph{Open-Source Instruction Fine-Tuned LLMs (Small)}}  \\ \midrule
LLaMA-2-7B-Chat~\cite{touvron2023llama} & 7B & OSS & 0.36 & 4.2 & 8.0 & 0.0 & 2.1 & 7.0 & 11.6 & - & - & - & - & - & - \\
Vicuna-7B-v1.5~\cite{chiang2023vicuna} & 7B & OSS & 0.43 & 9.7 & 8.7 & 0.0 & 2.5 & 9.0 & 2.2 & - & - & - & - & - & -\\
CodeLLaMA-7B-Instruct~\cite{roziere2023code} & 7B & OSS& 0.65 & 4.9 & 12.7 & 0.0 & 8.2 & 12.0 & 25.2 & - & - & - & - & - & - \\
CodeLLaMA-13B-Instruct~\cite{roziere2023code} & 13B & OSS & 0.74 & 3.5 & 9.7 & 0.0 & 10.4 & 14.0 & 43.8 & - & - & - & - & - & - \\
LLaMA-2-13B-Chat~\cite{touvron2023llama} & 13B & OSS & 0.66 & 4.2 & 11.7 & 6.0 & 3.6 & 13.0 & 25.3 & - & - & - & - & - & - \\
Vicuna-13B-v1.5~\cite{chiang2023vicuna} &13B & OSS & 0.86 & 10.4 & 6.7 & 8.0 & 9.4 & 12.0 & 41.7 & - & - & - & - & - & - \\
Groq-8B-Tool-Use~\cite{groq} & 8B & OSS & 1.27 & 15.3 & 11.7 & 4.0 & 17.6 & \underline{23.0} & 53.4 & 30.44 & 42.8 & 35.5 & 45.5 & 0.0 & \textbf{89.06} \\
LLaMA-3-8B-Instruct~\cite{dubey2024llama} & 8B & OSS & 1.51 & 18.1 & 12.3 & 24.0 & 15.9 & 19.0 & 56.1 & 35.79 & 60.6 & 66.2 & 48.4 & 0.5 & 59.57\\
LLaMA-3.1-8B-Instruct~\cite{dubey2024llama} & 8B & OSS & 1.74 & \underline{21.5} & 5.3 & \underline{34.0} & 18.4 & \textbf{25.0} & 59.5 & 46.76 & 70.3 & 76.5 & \underline{62.2} & 2.5 & 61.39\\
LLaMA-3-8B-IFT & 8B & OSS & \underline{2.07} & \textbf{22.2} & \underline{29.7} & 32.0 & \textbf{25.3} & 19.0 & \textbf{66.1}  & \underline{48.52} & \underline{72.5} & \underline{81.8} & \textbf{66.8} & \underline{2.6}  & 62.12\\
\rowcolor{teal!12} \method-8B-IFT & 8B & OSS & \textbf{2.29} & 20.8 & \textbf{41.7} & \textbf{46.0} & \underline{21.2} &17.0 & \underline{63.9}  & \textbf{50.59} & \textbf{84.3} & \textbf{86.2} & 60.1 & \textbf{9.6} & \underline{70.78}\\\midrule
\multicolumn{16}{l}{\emph{For Reference: Open-Source Instruction Fine-Tuned LLMs (Medium to Large) and API-based Commercial LLMs}}  \\ \midrule
LLaMA-2-70B-Chat~\cite{touvron2023llama} & 70B & OSS & 0.66 & 9.7 & 13.0 & 2.0 & 8.0 & 19.0 & 5.6 & - & - & - & - & - & - \\
CodeLLaMA-34B-Instruct~\cite{roziere2023code} & 34B & OSS & 1.13 & 2.8 & 14.0 & 4.0 & 23.5 & 20.0 & 52.1 & - & - & - & - & - & - \\
Gemini-1.5-Flash~\cite{reid2024gemini} & - & API &	1.81	& 20.1	& 46.0	& 22.0	& 14.2	& 17.0	& 39.1 & {53.01} & 77.1 & 71.2 & 71.2& 13.1 & 70.75\\
text-davinci-003~\cite{ouyang2022training} & - & API & 1.90 & 20.1 & 16.3 & 20.0 & 34.9 & 26.0 & 61.7 & - & - & - & - & - & - \\
DeepSeek-v2~\cite{liu2024deepseek} & 236B & OSS & 1.97	& 20.8	& 21.7	& 38.0	& 21.7	& 22.0	& 57.4 & - & - & - & - & - & -\\
Mixtral-8x22B~\cite{jiang2024mixtral} & 176B & OSS	& 2.00	& 24.3	& 25.7	& 14.0	& 31.1	& 28.0	& 62.8 & 43.00 & 56.1 & 59.7 & 65.3 & 8.9 & 63.26\\
gpt-3.5-turbo-0125~\cite{chatgpt} & - & API & 2.12 &	32.6	&	36.7	&	16.0	&	25.9	&	20.0	&	{64.1} & 51.90 & 84.5 & 81.7 & 59.0 & {19.1} & 66.53 \\
Claude-3-Haiku~\cite{claude-3} & - & API & 2.13	& 14.6	& 41.0	& 42.0	& 27.3	& 14.0	& 57.8 & 38.39 & 62.6 & 60.7 & 58.1 & 1.6 & 55.47 \\
Command-R-Plus-FC~\cite{command-r-plus} & - & API & - & - & - & - & - & - & - & 45.22 & 77.7 & 77.4 & 54.2 & 6.1 & 76.29\\
LLaMA-3-70B-Instruct~\cite{dubey2024llama} & 70B & OSS & 2.73 & 28.6 & {50.3}	& 44.0	& 39.5	& 22.0	& 53.6 & 49.55 & {87.2} & {87.4} & {63.4} & 1.1 & 84.95\\
gpt-4-0613~\cite{achiam2023gpt} & - & API & {4.52} & {42.4} & 32.0 & {78.0} & {58.8} & {29.0} & 61.1 & - & - & - & - & - & {89.26} \\\bottomrule
\end{tabular}
\caption{Main experiments on three agent benchmarks across various model scales. 
\textbf{Bold} and \underline{underlined} texts represent the best and the second-best results, respectively. Notations are consistent throughout all tables. ``OSS'', ``API'', and ``OA'' denote ``Open-Sourced LLMs'', ``API-based Commercial LLMs'', and ``Overall'', respectively.
}\label{tab:agentbench}
\vspace{-1ex}
\end{table*}

\subsection{Main Experiments: \method-8B-Base}
Following~\citet{shao2024deepseekmath,dubey2024llama}, we evaluate our two-stage pre-trained \texttt{\method-8B-Base} on three agent-specific benchmarks (API-Bank, API-Bench, NexusRaven) and one general benchmark (MMLU).
We observe that incorporating more agent data during pre-training consistently reduces benchmark loss on agent tasks in Figure \ref{fig:val_loss} (b). Additionally, Figure \ref{fig:val_loss} (c) demonstrates that \texttt{\method-8B-Base} achieves significantly lower benchmark loss compared to the \texttt{LLaMA-3-8B} series of base models.
Furthermore, Table \ref{tab:agentbench} reports the benchmark scores, where \texttt{\method-8B-Base} leads in performance across all benchmarks among the open-source base models. 
Our findings indicate that both pre-training stages (I \& II) enhance \method's fundamental capabilities across a wide range of agent tasks without compromising general capabilities.

\subsection{Main Experiments: \method-8B-IFT}
Table \ref{tab:agentbench} presents the main experimental results of instruction fine-tuned \method and baselines. 
\method consistently outperforms small to medium size open-source LLMs. Moreover, \texttt{\method-8B-IFT} remains competitive compared to baseline models with significantly more parameters or commercial LLMs. 

\noindent \textbf{Enhanced Capabilities Through Pre-training.}
We conduct a direct comparison between \method and \texttt{LLaMA-3-8B-Base}~\cite{dubey2024llama}, both instruction-tuned using the same instruction fine-tuning data. \texttt{\method-8B-IFT} outperforms \texttt{LLaMA-3-8B-IFT} across all three benchmarks, indicating that the observed improvements can be attributed to the pre-training stage. 
Moreover, incorporating more domain-specific knowledge during the pre-training stage leads to better performance, without requiring additional instruction fine-tuning data.

\noindent \textbf{Excelling in Complex Multi-turn Tasks.}
BFCL-v3, the latest benchmark, emphasizes multi-turn tool function-calling tasks requiring intrinsic reasoning capabilities and function-calling proficiency. 
Due to its recent introduction, the limited availability of task-specific data for instruction-tuning has led to suboptimal performance, particularly in multi-turn function-calling accuracy, as observed with models like Groq-8B-Tool-Use \cite{groq}. 
In contrast, \method exhibits significantly better performance on BFCL-v3, suggesting that its improvements in core agentic capabilities and generalization stem from pre-training on our large-scale, diverse agent-oriented corpus.

\begin{table}[t]
\centering
\fontsize{7}{9}\selectfont\setlength{\tabcolsep}{0.3em}
\begin{tabular}{@{}l>{}ccccccc>{}c@{}}
\toprule
\textbf{Datasets ($\rightarrow$)} & \multicolumn{7}{c}{\textbf{AgentBench}} & \textbf{BFCL-v2}\\
\cmidrule(lr){2-8} \cmidrule(lr){9-9}
\textbf{Models ($\downarrow$)} & \textbf{OA} & \textbf{OS} & \textbf{DB} &  \textbf{HH} & \textbf{KG} & \textbf{WB} & \textbf{WS} & \textbf{OA}\\\midrule
\rowcolor{teal!12} \method-8B-Base & \textbf{1.87} & 20.8 & 32.3 & 30.0 & 16.0 & \textbf{16.0} & 60.5 & \textbf{25.18}\\
\quad w/ Stage-1 PT Only & 1.76	& 20.1	& 29.0	& 28.0	& \textbf{17.5}	& 14.0	& 56.1 & 23.88 \\\midrule
\quad w/o Data Filtering & 1.85	& 20.8	& \textbf{36.3}	& 28.0	& \textbf{17.5}	& 14.0	& 54.0 & 21.08 \\
\quad w/o Retrieval Data & 1.84	& \textbf{22.9}	& 16.7	& \textbf{48.0}	& 5.4	& \textbf{16.0}	& \textbf{66.4}  & 19.35\\\midrule
\rowcolor{teal!12} \method-8B-IFT & \textbf{2.29} & 20.8 & \textbf{41.7} & \textbf{46.0} & \textbf{21.2} &17.0 & 63.9  & \textbf{70.78}\\
\quad w/ Stage-1 PT Only & 2.00	& 20.8	& \textbf{41.7}	& 34.0	& 18.4	& 10.0	& 63.9
& 64.23 \\\midrule
\quad w/o Data Filtering & 2.10 &	\textbf{23.6} & 28.3	& 44.0	& 17.2	& \textbf{18.0}	& \textbf{64.0} & 59.34 \\
\quad w/o Retrieval Data & 1.99	& 21.5	& 30.3	& 38.0	& 17.2	& 17.0	& 60.7 & 49.86\\\bottomrule
\end{tabular}
\caption{Ablation studies on the effect of (1) different pre-training stages and (2) retrieved data.
}\label{tab:ablation}
\vspace{-3ex}
\end{table}

\subsection{Ablation Studies}
Table~\ref{tab:ablation} presents the ablation results of \method on AgentBench and BFCL-v2. 

\noindent\textbf{Effect of Pre-Training Stages.}
Removing the second pre-training stage results in a slight performance decline for both base and instruction-tuned models across all tasks. Although the Stage-I pre-training data, comprising a large volume of general and retrieved agent data from the web, brings the \dataset closer to the general data distribution, it still differs from the data used in downstream applications and evaluations. The Stage-II pre-training is essential for effectively bridging the gap between the pre-training corpus and the instruction fine-tuning data, thereby enhancing overall model performance.

\noindent\textbf{Effect of Retrieved Data.}
Degrading the retrieved data to unfiltered, low-quality data or removing it entirely negatively impacts overall performance.
For tasks with numerous hand-crafted instructions and simulated trajectories available on the open web (\eg, HH and WS), the seed data of \dataset can lead to model overfitting on specific patterns.
When the large volume of retrieval data is removed, the seed data predominates, leading to improved performance on these specific tasks but reduced performance on others.

\begin{table}[t]
\centering
\fontsize{7}{9}\selectfont\setlength{\tabcolsep}{0.3em}
\begin{tabular}{@{}l>{}ccc}
\toprule
\textbf{Models ($\downarrow$)$\slash$Datasets ($\rightarrow$)} & \textbf{AgentBench} & \textbf{BFCL-v3} & \textbf{BFCL-v2}\\\midrule
\rowcolor{teal!12} \method-8B-IFT & \textbf{2.29} & \textbf{51.59}  & \textbf{70.78}\\
LLaMA-3-8B-IFT & 2.07 \tiny{\textcolor{red}{(-9.6\%)}} & 48.52 \tiny{\textcolor{red}{(-6.0\%)}} & 62.12 \tiny{\textcolor{red}{(-12.2\%)}}\\\midrule
Groq-8B-Tool-Use~\cite{groq} & 1.27 \tiny{\textcolor{red}{(-44.5\%)}} & 30.44 \tiny{\textcolor{red}{(-41.0\%)}} & 89.06 \tiny{\textcolor{green}{(+25.8\%)}} \\
AgentLM-7B~\cite{zeng2023agenttuning} & 2.36 \tiny{\textcolor{green}{(+3.1\%)}} & 16.67 \tiny{\textcolor{red}{(-67.7\%)}} & 19.18 \tiny{\textcolor{red}{(-72.9\%)}}\\
ToolACE-8B~\cite{liu2024toolace} & 1.48 \tiny{\textcolor{red}{(-35.3\%)}} & 58.20 \tiny{\textcolor{green}{(+12.8\%)}} & 91.41 \tiny{\textcolor{green}{(+29.2\%)}} \\\bottomrule
\end{tabular}
\caption{Generalization across three agent benchmarks.
}\label{tab:generalization}
\vspace{-3ex}
\end{table}

\begin{table*}[t]
\centering
\fontsize{7}{9}\selectfont\setlength{\tabcolsep}{0.3em}
\begin{tabular}{@{}l>{}ccccc>{}ccccc@{}}
\toprule
\textbf{} & \multicolumn{5}{c}{\textbf{Benchmark Metrics ($\uparrow$)}} & \multicolumn{5}{c}{\textbf{Benchmark Loss ($\downarrow$)}}\\
\cmidrule(lr){2-6} \cmidrule(lr){7-11}
\textbf{Models ($\downarrow$) / Datasets ($\rightarrow$)} & \textbf{GSM8K}  & \textbf{HumanEval} & \textbf{HumanEval+} &  \textbf{BBH} & \textbf{OA}  & \textbf{IFEval} & \textbf{hellaswag}  & \textbf{MMLU}  &  \textbf{BBH}  & \textbf{OA}  \\\midrule
LLaMA-3-8B~\cite{dubey2024llama}           & 0.420 & 0.372 & 0.317 & 0.613  & 0.431 & {0.648} & {0.759} & {0.526} & {0.361} & {0.573} \\
\rowcolor{teal!12} \method-8B-Base       & 0.460 & 0.411 & 0.356 & 0.584  & 0.453 & 0.683 & 0.769 & 0.536 & 0.374 & 0.591 \\ \midrule
LLaMA-3-8B-IFT       & {0.695} & 0.343 & 0.337 & {0.596}  & 0.493 & 1.046 & 0.908 & 0.725 & 0.503 & 0.795 \\
\rowcolor{teal!12} \method-8B-IFT    & 0.686 & {0.373} & {0.373} & 0.567 & {0.500} & {0.657} & {0.784} & {0.559} & {0.369} & {0.592} \\ \midrule
ToolACE-8B~\cite{liu2024toolace}           & 0.623 & 0.385 & 0.324 & 0.120  & 0.363 & 0.774 & 0.848 & 0.602 & 0.442 & 0.666 \\
AgentLM-7B~\cite{zeng2023agenttuning}           & 0.549 & 0.122 & 0.110 & 0.071  & 0.213 & 0.783 & 0.915 & 0.657 & 0.450 & 0.701 \\
LLaMA-3-8B-Instruct~\cite{dubey2024llama}  & 0.797 & 0.646 & 0.573 & 0.660  & 0.669 & 0.619 & 0.769 & 0.533 & 0.361 & 0.570 \\
\bottomrule
\end{tabular}
\caption{Comprehensive evaluation of general model capabilities across diverse benchmarks. \method maintains general capabilities while achieving competitive performance against baseline and specialized models. 
}\label{tab:preserve}
\vspace{-3ex}
\end{table*}

\subsection{Cross-Task Generalization}
Table~\ref{tab:generalization} compares \method with several instruction fine-tuned agent frameworks across three agent benchmarks for cross-task generalization.
While models fine-tuned on task-specific data excel in corresponding tasks~\cite{groq,zeng2023agenttuning,liu2024toolace}, they struggle to generalize across different agent benchmarks.
In contrast, \method performs consistently well across all tasks, suggesting that the large and diverse pre-training corpora, \dataset, effectively enhance function calling and agentic reasoning, leading to better generalization. 
Furthermore, the compared methods are based on continued instruction fine-tuning of \texttt{LLaMA-3-8B-Instruct}, which inherently possesses strong instruction-following and understanding capabilities due to its meticulously curated post-training data. 
Unlike models relying solely on instruction fine-tuning, the pre-training of \method effectively improves its fundamental capabilities, thereby offering a more robust foundation for diverse agentic applications.

\subsection{Preservation of General Capabilities}
To evaluate the preservation of general capabilities, we further conduct comprehensive experiments across seven additional benchmarks (Table~\ref{tab:preserve}) besides MMLU, spanning mathematics~\cite{cobbe2021training}, software development~\cite{chen2021evaluating,liu2024your}, logical reasoning~\cite{suzgun2022challenging}, and broad language model abilities~\cite{zhou2023instruction,zellers2019hellaswag,hendrycks2020measuring}. Our results demonstrate that \method maintains comparable performance to the base model across these diverse domains while significantly enhancing agent-specific capabilities.

\section{Conclusion}
\label{sec:conclusion}

In summary, \dataset and \method collectively advance open-source LLM-based autonomous agents by addressing critical gaps in pre-training corpora. 
Through exhaustive scaling law experiments, we identify an empirically optimal data mix ratio of approximately 1:1:1 for agent, code, and text data, maximizing the fundamental and generalization capabilities of LLM agents. 
Empirical evaluations underscore the efficacy and validity of \dataset in fostering enhanced fundamental agentic capabilities and superior generalization in LLM-based autonomous agents.



\section*{Limitations}
\noindent \textbf{Data Composition.} 
While knowledge of the composition of pre-training or instruction fine-tuning data would further enhance the effectiveness of \method, most prominent open-source LLMs (\eg, \texttt{LLaMA-3-8B-Instruct}) do not disclose detailed data information. Nevertheless, our continual pre-training experiments with \texttt{LLaMA-3-8B} demonstrate that significant improvements are achievable even without this knowledge. 

\noindent \textbf{Model Scalability.} Computational constraints currently restrict our ability to extend these experiments to larger models. In future work, we aim to validate our findings and methodologies on more expansive LLM architectures, pending access to increased computational resources.


\section*{Ethical Statement}
\noindent \textbf{Data Contamination.} 
A potential concern in our evaluations is test set contamination, which occurs when some task-specific examples overlap with data used during continual pre-training~\cite{oren2024proving}. To mitigate this issue, we follow~\citet{wang-etal-2024-improving-text} and conduct a string-matching analysis, which indicates no overlap between our training data and the datasets of the target tasks. Moreover, we intentionally exclude all evaluation benchmark data from both our pre-training and fine-tuning datasets to ensure a fair comparison.

\noindent \textbf{Reproducibility.}
To promote transparency, reproducibility, and generalizability in our research, we include all details of the dataset construction (\eg, data collection, processing, retrieving, filtering, scaling law, \etc) of \dataset in \cref{sec:dataset} and the training procedures for \method in \cref{sec:method}. 
Experimental setups and results are presented in \cref {sec:exp}.
Additionally, we detail the pre-training, instruction fine-tuning, and testing tasks and datasets in \cref{app:data-pretrain,app:data-ift,app:data}, respectively. 


\bibliography{ref}

\appendix
\onecolumn
\begin{center}
	{\Large \textbf{Supplementary Materials for \method}}
\end{center}

\newenvironment{titledframe}[1]
  {\mdfsetup{
    frametitle={\colorbox{white}{\space#1\space}},
    innertopmargin=10pt,
    frametitleaboveskip=-\ht\strutbox,
    frametitlealignment=\center
    }
  \begin{mdframed}%
  }
  {\end{mdframed}}

\startcontents[sections]
\printcontents[sections]{l}{1}{\setcounter{tocdepth}{2}}

\newpage
\section{Task and Dataset Information}
\label{app:task}

\subsection{Pre-Training Corpus: \dataset}
\label{app:data-pretrain}
\begin{figure}[ht]
  \centering
  \includegraphics[width=0.98\linewidth]{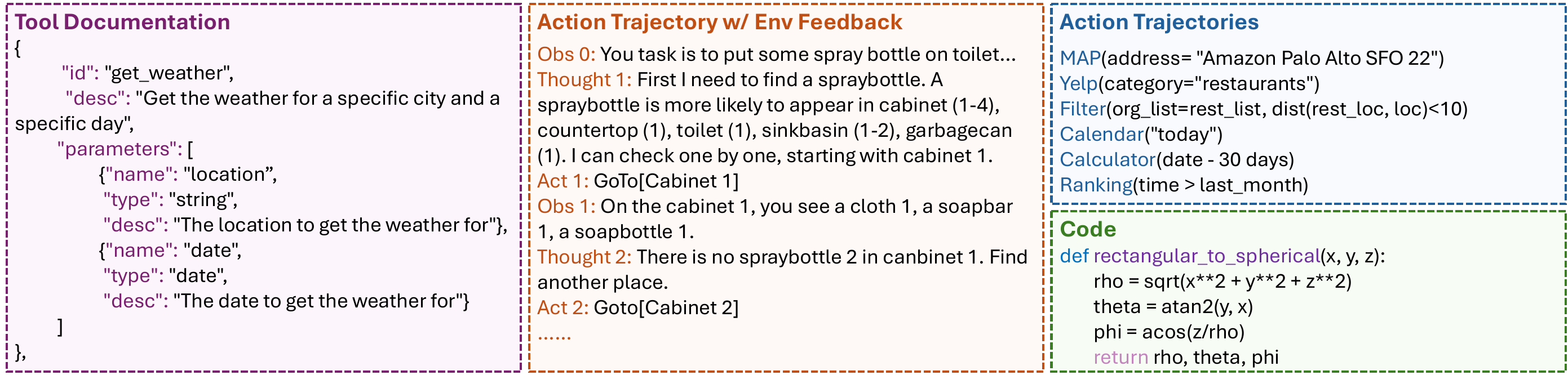}
  \caption{
  Examples of different task formats in \dataset, including tool documentation, action trajectory (w/ environmental feedback), and code data. 
  }
  \label{fig:datasample}
\end{figure}

\noindent \textbf{Agent Data Sources.}
To promote transparency, reproducibility, and potential generalization to novel domains in agent research, we publicly release the training recipe utilized for \dataset during the pre-training stage. To enhance the fundamental capabilities of \method, we compile a unique, comprehensive, and large-scale corpus of agent data sources, including API documentation, API function calling trajectories, code, and text data. Tables~\ref{tab:data-source1} and \ref{tab:data-source2} provide a comprehensive overview of \dataset used in \method, detailing the data sources, their respective sizes, and public availability status. 
All data sources utilized in \dataset are licensed under \texttt{Apache-2.0}, \texttt{MIT}, or \texttt{LGPL-2.1}, permitting non-commercial use and aligning with the research objectives of this work.
Examples of task formats in \dataset are available in Figure~\ref{fig:datasample}.

\noindent \textbf{Text and Code Data.}
Since agent data typically includes detailed task descriptions, formatted function calls, and environmental feedback, significant gaps exist between agent data and standard text and code data. Given that current open-sourced LLMs have already been pre-trained on text and code data, and to preserve their generalization ability, it is necessary to mix agent data with text and code data during the continual pre-training stage. For the text data, we primarily select a corpus that covers commonsense reasoning, mathematical reasoning, scientific reasoning, and general text.

\noindent$\bullet$ \textbf{RedPajama\_CommonCrawls}\footnote{\url{https://www.together.ai/blog/redpajama-data-v2}}~\cite{raffel2020exploring} is a large-scale web text dataset collected by the RedPajama project. It encompasses a diverse range of internet texts, including blogs, news articles, forum discussions, and social media posts. Incorporating this dataset helps to preserve general language understanding and generation capabilities, as it captures a wide variety of writing styles and topics, thus offering significant linguistic diversity.

\noindent$\bullet$ \textbf{Encyclopedic Content} is a comprehensive knowledge base sourced from Wikipedia\footnote{\url{https://www.wikipedia.org/}} and WikiQA~\cite{yang-etal-2015-wikiqa}. This dataset includes extensively curated articles covering a wide range of human knowledge domains. Incorporating encyclopedic content during continual pre-training helps ensure factual accuracy and reliability in the model's learned information.

\noindent$\bullet$ \textbf{Textbooks} from OpenStax\footnote{\url{https://openstax.org/}} provide peer-reviewed, openly licensed textbooks for higher education. These textbooks span topics such as mathematics, science, economics, and the humanities. Since textbooks are structured with well-organized chapters and summaries, continual pre-training on this corpus exposes the model to formal educational language and coherent knowledge representation.

\noindent$\bullet$ \textbf{Mathematical Content} from OpenWebMath~\cite{paster2024openwebmath} aggregates open-access mathematical texts, problem sets, and explanations. This dataset spans topics ranging from pure mathematics to applied fields, enabling the model to understand and generate mathematically rigorous content.

\noindent$\bullet$ \textbf{arXiv Papers}\footnote{\url{arxiv.org}} include preprints hosted on arXiv in fields such as physics, mathematics, computer science, and more. This dataset features advanced terminology, methodologies, and academic discourse. Using this data for continual pre-training enhances the model's ability to grasp complex scientific concepts and fosters cross-disciplinary understanding.

\noindent$\bullet$ \textbf{StarCoder-v2}~\cite{lozhkov2024starcoder} is a large-scale collection of source code curated to advance research in code generation and understanding. We select all documentation samples and randomly sample the remaining portion for inclusion in the \dataset. This dataset provides knowledge of complex programming patterns and semantics, which may benefit the tool-function-calling capabilities of LLMs.

\subsection{Instruction Fine-Tuning Task and Dataset}
\label{app:data-ift}
The instruction fine-tuning stage empowers LLMs with instruction-following capabilities and aligns LLM agents with task-specific requirements and user preferences. To facilitate direct and fair comparison, we employ a diverse range of tasks for both the instruction fine-tuning baseline model, \texttt{LLaMA-3-8B-IFT}, and our model, \texttt{\method-8B-IFT}, including {(1) a general conversation dataset,} \emph{ShareGPT}~\cite{chiang2023vicuna}; {(b) a single-tool function-calling conversation dataset,} \emph{ToolACE}~\cite{liu2024toolace}; and
{(c) a multi-turn planning conversation dataset,} \emph{AgentFlan}~\cite{chen2024agent}.

\noindent $\bullet$ \textbf{ShareGPT}~\cite{chiang2023vicuna} is a general dataset comprising real-world conversations from 70K user data, designed to fine-tune models for enhanced instruction-following capabilities. It significantly improves LLMs' ability to handle complex, multi-turn dialogues. The dataset encompasses a wide range of topics and natural, human-generated prompts, enabling models to learn from authentic interactions. By leveraging real user data, ShareGPT allows models to better generalize across diverse tasks and navigate increasingly complex instructions, closely mimicking real-world conversational scenarios.

\noindent $\bullet$ \textbf{ToolACE}~\cite{liu2024toolace} is a single-tool conversation dataset designed to enhance the function-calling capabilities of LLM agents. It comprises 26,507 APIs across 30 primary domains (\eg, entertainment) and is categorized into 390 coarse-grained sub-domains (\eg, music). In addition, ToolACE accommodates complex nested parameters, manages both parallel and dependent function calls, and encompasses a wide variety of tool-related data.

\noindent $\bullet$ \textbf{AgentFlan}~\cite{chen2024agent} is a multi-turn planning dataset that combines data in two formats: 10\% in ReAct format and 90\% in conversation format. It encompasses 24,703 instances derived from AgentInstruct and ToolBench. AgentFlan deliberately excludes format-following instructions and common reasoning tasks from its training corpus, aiming to elicit pure agent abilities from LLMs without overfitting to specific format protocols.

\subsection{Evaluation Task and Dataset}
\label{app:data}
We conduct the main experiments of \method on three widely used LLM agent benchmarks across a wide range of scenarios, including: 

\noindent $\bullet$ \textbf{AgentBench}~\cite{liu2024agentbench} presents six distinct environments in a multi-turn, open-ended generation setting: Operating System (OS), Database (DB), Knowledge Graph (KG), House-Holding (HH), Web Shopping (WS), and Web Browsing (WB). We leverage AgentBench to evaluate intrinsic reasoning and adaptation to environmental feedback.

\noindent $\bullet$ \textbf{Berkeley Function Calling Leaderboard (BFCL)}~\cite{patil2023gorilla} provides a rigorous framework for assessing the function-calling proficiencies of diverse LLM agents. This benchmark encompasses 2,000 question-function-answer triads, spanning multiple programming paradigms (Python, Java, JavaScript, REST API) and heterogeneous application domains. The BFCL's evaluation protocol incorporates varying degrees of complexity, ranging from single-function selection tasks to scenarios necessitating the concurrent execution of multiple-function calls. Notably, the latest iteration, BFCL-v3, represents a significant methodological advancement over its predecessor by introducing a novel category that evaluates multi-turn and multi-step function invocation, more closely simulating real-world tool usage scenarios. We leverage BFCL-v2 and -v3 to evaluate the function-calling capability of LLM agents. 

Following~\citet{dubey2024llama}, we leverage additional three agent benchmarks (Nexus~\cite{srinivasan2023nexusraven}, API-Bank~\cite{li2023api}, and API-Bench~\cite{patil2023gorilla}) and one general benchmark (MMLU)~\cite{hendrycks2020measuring} for benchmark loss in the scaling law experiments.

\section{Baseline Details}
\label{app:baseline}

\subsection{Base LLMs}

\noindent $\bullet$ \textbf{\texttt{LLaMA-3-8B-Base}}~\cite{dubey2024llama} is a small-scale flagship model in Meta's LLaMA-3 series, featuring 8 billion parameters. We compare \method with \texttt{LLaMA-3-8B-Base}, which also serves as the backbone of \texttt{\method-8B-Base}, to demonstrate the effectiveness of continual pre-training.

\noindent $\bullet$ \textbf{\texttt{LLaMA-3.1-8B-Base}}~\cite{dubey2024llama} is an improved version of \texttt{LLaMA-3-8B}, offering more efficient parameter utilization and enhanced fine-tuning capabilities. The 3.1 series models are optimized for multilingual support and scalability, allowing for a longer context length of up to 128K tokens. We select \texttt{LLaMA-3.1-8B-Base} as the current state-of-the-art small-scale open-sourced base model for comparison.

\subsection{Open-Source Instruction Fine-tuned LLMs}
We compare \texttt{\method-IFT} with the following open-sourced instruction-tuned LLMs:

\noindent$\bullet$ \textbf{\texttt{LLaMA-2-Chat}}~\cite{touvron2023llama} is a series of large language models developed by Meta, designed for conversational AI. The models support text-based interactions and come in varying parameter sizes, such as 7B, 13B, and 70B. For comparison, we select models of comparable scale, specifically \texttt{LLaMA-2-7B-Chat} and \texttt{LLaMA-2-70B-Chat}.

\noindent$\bullet$ \textbf{\texttt{Vicuna-v1.5}}~\cite{chiang2023vicuna} is a collection of open-source LLMs fine-tuned from LLaMA models, optimized for high-quality conversational abilities. These models are fine-tuned using datasets derived from user-shared conversations and are available in sizes such as 7B and 13B parameters, both of which are included in our comparisons.

\noindent$\bullet$ \textbf{\texttt{CodeLLaMA}}~\cite{roziere2023code} is a specialized extension of the LLaMA family designed for code generation and understanding. Built upon LLaMA-2, CodeLLaMA introduces enhancements tailored to coding tasks. We evaluate multiple sizes, including \texttt{CodeLLaMA-7B-Instruct}, \texttt{CodeLLaMA-13B-Instruct}, and \texttt{CodeLLaMA-34B-Instruct}.

\noindent$\bullet$ \textbf{\texttt{Groq-8B-Tool-Use}}~\cite{groq} is a specialized variant of LLaMA-3-8B, fine-tuned by Groq for advanced tool use and function-calling tasks. It leverages post-training techniques to achieve state-of-the-art performance in function-calling tasks, including BFCL.

\noindent$\bullet$ \textbf{\texttt{LLaMA-3-Instruct}}~\cite{dubey2024llama} belongs to Meta's LLaMA-3 family, optimized for instruction-following tasks. These models excel at tasks requiring explicit instructions, making them suitable for applications such as chatbots, virtual assistants, and task-specific text generation. We compare \texttt{LLaMA-3-8B-Instruct} and \texttt{LLaMA-3.1-8B-Instruct} as small-scale state-of-the-art instruction-tuned models. Additionally, we use \texttt{LLaMA-3-70B-Instruct} as a reference model for comparison.

\noindent$\bullet$ \textbf{\texttt{DeepSeek-v2}}~\cite{liu2024deepseek} and \textbf{\texttt{Mixtral-8x22B}}~\cite{jiang2024mixtral} are both cutting-edge language models utilizing Mixture-of-Experts (MoE) architectures to optimize efficiency and performance across various domains. We include both models as reference points in our comparisons.

\subsection{API-based Commercial LLMs (for reference)}
We also consider {API-based commercial LLMs} for reference only, including \textbf{\texttt{Gemini-1.5-Flash}}~\cite{reid2024gemini}, \textbf{\texttt{text-davinci-003}}~\cite{ouyang2022training}, \textbf{\texttt{gpt-3.5-turbo-0125}}~\cite{chatgpt}, \textbf{\texttt{gpt-4-0613}}~\cite{achiam2023gpt}, \textbf{\texttt{Claude-3-Haiku}}~\cite{claude-3}, and \textbf{\texttt{Command-R-Plus-FC}}~\cite{command-r-plus}.
We exclude prompting and instruction fine-tuned agent frameworks from our main experiments to focus on evaluating the fundamental agentic capabilities of LLMs.







\section{Additional Related Works}
\label{app:related works}
LLM-based intelligent agents and autonomous entities have demonstrated proficiency in tool utilization~\cite{qin2023toolllm,zhuang2024toolqa}, decision-making~\cite{wang2023voyager,li2024matryoshka}, and action execution through interactions with diverse environments~\cite{sun2024adaplanner,shi2024ehragent}.

\subsection{Black-box LLM Agents}
Existing methods for enhancing commercial closed-source LLM-based agents primarily focus on designing task-specific prompts. These prompts often incorporate tool function documentation \citep{hsieh2023tool}, few-shot demonstrations \citep{lu2024chameleon}, environmental feedback \citep{yao2022react, sun2024adaplanner, wang2023voyager}, and tree-like reasoning procedures \citep{yao2024tree, zhuang2023toolchain}. While these approaches have yielded improved results and increased flexibility, they come with significant drawbacks. The use of closed-source LLMs incurs substantial financial costs and raises safety concerns \citep{li2023multi,zhuang2024hydra, yuan2023gpt,sun2024bbox,shi2024medadapter}, limiting their wider deployment. Moreover, these prompting techniques do not fundamentally enhance the inherent agent abilities of the LLMs. Instead, they rely heavily on the function-calling capabilities of closed-source LLMs, which may lack stability across different updates or versions\footnote{\url{https://openai.com/index/function-calling-and-other-api-updates/}}.

\subsection{White-box LLM Agents}
Open-source LLMs have recently emerged as promising alternatives, demonstrating effectiveness in various applications \citep{touvron2023llama, jiang2024mixtral,tang-etal-2024-mimir}. While these models excel in natural language processing tasks, they still underperform when serving as the core of LLM agents \citep{zeng2023agenttuning, liu2024agentbench}. This limitation is primarily due to insufficient training samples and smaller model scales compared to their closed-source counterparts.
Researchers have attempted to address these shortcomings through various approaches. Some have fine-tuned LLMs with specific API documentation and function call sequences \citep{qin2023toolllm, gou2024tora}. Others have leveraged domain-specific data to learn tool embeddings or modify the decoding process \citep{schick2024toolformer, hao2024toolkengpt, zhang2023syntax}. However, this focus on specialized capabilities often comes at the expense of the LLMs' general abilities and compromises their generalizability.
A recent approach by \citet{chen2024agent} attempts to mitigate this issue by composing API function sequential data from diverse sources and reorganizing the training corpus. Yet, compared to the breadth of data included in the pre-training stage, the collected data from five to six different sources represents only a small fraction of real-world decision-making scenarios, limiting generalization to new tasks.
Moreover, the superficial alignment hypothesis \citep{zhou2024lima} suggests that a model's fundamental knowledge and capabilities are acquired almost entirely during pre-training. Post-training techniques merely guide the model in selecting which subdistribution of formats to use when interacting with users. Consequently, core abilities cannot be significantly improved through prompting and post-training techniques alone.

\subsection{Finetuning-based LLM Agents} 
Table \ref{tab:related} summarizes existing instruction fine-tuning-based LLM agents and their training samples.
For example, Gorilla \citep{patil2023gorilla} fine-tuned a LLaMA-based model using API documentation and demonstrations from Huggingface, TorchHub, and TensorFlowHub. Toolformer \citep{schick2024toolformer} introduced special tokens around API function calls to teach the model when and how to leverage tools during fine-tuning. ToolkenGPT \citep{hao2024toolkengpt} incorporated tools as special tokens into the model's vocabulary, while ToolLLaMA \citep{qin2023toolllm} built datasets rich in various tools.
However, these methods often rely on APIs and datasets from similar domains, potentially limiting their effectiveness to tasks within those domains. To address this limitation, recent instruction tuning methods \citep{achiam2023gpt, srinivasan2023nexusraven, zeng2023agenttuning, chen2024agent} have expanded to include a diverse range of API function call data and tasks, aiming to equip models with broader generalization capabilities across different planning tasks.
Nevertheless, the superficial alignment hypothesis \citep{zhou2024lima} suggests that a model's fundamental knowledge and capabilities are predominantly acquired during pre-training. According to this hypothesis, post-training techniques such as instruction tuning and alignment primarily teach the model which sub-distributions of formats to utilize when interacting with users, rather than fundamentally expanding its capabilities.
Moreover, heavy fine-tuning prevents generalization and degrades performance in general use cases, potentially suppressing the original base model capabilities~\cite{ghosh2024a}.

\subsection{Pretraining-based LLM Agents} 

To overcome the limitations of prompting and tuning-based methods, recent initiatives have focused on pre-training or continual pre-training of language models to bolster their fundamental capabilities. Several notable examples have emerged in this domain:
CodeGen \citep{nijkamp2023codegen} and CodeLLaMA \citep{roziere2023code} enhance the coding skills of LLMs. Building on the success of these code LLMs, LEMUR \citep{xu2024lemur} further instruction tunes a code LLM with additional assistant and tool-related data.
Pandora \citep{xiang2024pandora} represents a pre-trained world model that incorporates visual encoders to process a wide array of multi-modal data, including videos and textual actions.
The most closely related work to our proposed model is OpenFunctions-v2 \citep{patil2023gorilla}. This model is pre-trained on a vast collection of data sources, including 19,353 Python packages, 16,586 Java repositories, 4,285 JavaScript repositories, 6,009 public APIs, and 19,090 command line tools. However, while OpenFunctions-v2 primarily focuses on making correct API function calls, it lacks emphasis on the intrinsic reasoning abilities required for managing multiple API function calls, as well as adapting to environmental feedback.
\section{Dataset Construction Details}
\label{app:datacollection}
\begin{figure}[ht]
  \centering
  \includegraphics[width=0.86\linewidth]{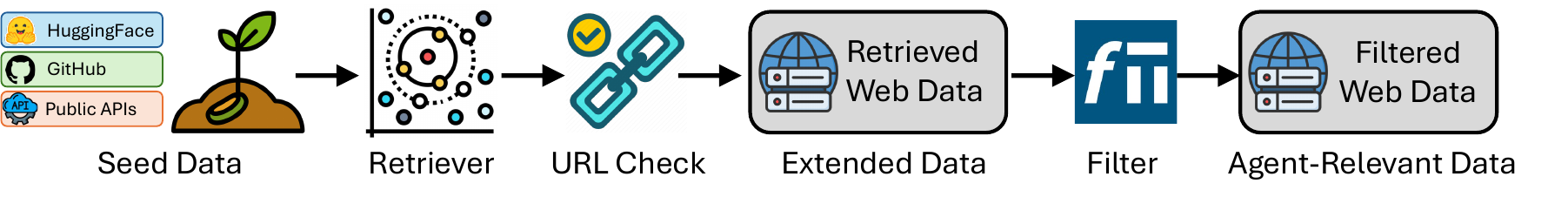}
  \caption{Overview of the data collection workflow in \dataset.  }
  \label{fig:dataprepare}
\end{figure}

To scale and diversify the pre-training corpus for LLM agents, we introduce a three-stage construction process (Figure~\ref{fig:dataprepare}) for \dataset in \cref{sec:dataset}. We then include additional data collection details as follows.

\subsection{Seed Data Collection Details}
\label{app:4-1}

We begin by assembling a set of high-quality initial data samples to establish a robust foundation.
Specifically, we systematically explore publicly accessible resources to gather high-quality API documentation and associated action trajectories. This includes compiling a diverse dataset of agent behavior from public repositories, official API documentation sources, and data synthesized through LLMs.
Given that the volume of tool-related data remains significantly smaller than that of plain text or code data, we employ data augmentation and generation techniques to expand the tool-related dataset.

\subsubsection{Public APIs.}
First, we collect data from over 1,400 public API documentations\footnote{\url{https://github.com/public-apis/public-apis}} and integrate additional data from official websites, including Huggingface\footnote{\url{https://huggingface.co/docs}}, TorchHub\footnote{\url{https://pytorch.org/docs/stable/index.html}}, and Python Modules\footnote{\url{https://docs.python.org/3/index.html}}, among others.
This compilation includes detailed API definitions and parameter descriptions, enabling the model to gain a better understanding of API functions.
As the depth and location of the documentation vary across different API websites, we apply a three-level scraping strategy: 
(1) \emph{Level 1:} the collected 1,400 URLs; 
(2) \emph{Level 2:} 37,753 URLs appearing on the Level 1 web pages; 
(3) \emph{Level 3:} 83,468 URLs appearing in the Level 2 web pages. 
We then apply URL checks to verify validity and filter for documentation-relevant data by searching for keywords (\eg, ``doc'', ``guide'', ``reference'', \etc).

\subsubsection{Public Repositories.}
To strengthen the model's intrinsic reasoning and planning abilities, we integrate publicly available action trajectories from over $60$ public repositories of related papers and datasets. These action trajectories span multiple domains, including programming code, natural language reasoning steps, embodied AI action sequences, grounded multi-modal data, web interactions, and function call sequences. This diverse range of trajectories, incorporated during the pre-training phase, enhances the model's reasoning capabilities and improves its generalization to various scenarios.

\subsubsection{Code-to-Text Synthesis.}
Given the limited quantity and API coverage of curated data from public APIs and repositories, we exploit the strong generative abilities of LLMs to synthesize additional API documentation and use cases. To produce high-quality synthetic agent data, we utilize \emph{StarCoder-API}\footnote{\url{https://huggingface.co/datasets/luna-code/starcoderdata-apis}} as a knowledge base, which includes code snippets involving third-party APIs. Based on these code snippets and the API function calls within them, we generate corresponding API documentation and associated use cases.
For efficiency, we utilize multiple LLMs from Amazon Bedrock\footnote{\url{https://aws.amazon.com/bedrock/}} for data synthesis, including \texttt{Claude-3-Sonnet}, \texttt{Claude-3-Haiku}~\cite{claude-3}, \texttt{Mistral-Large}~\cite{mistral-large}, \texttt{LLaMA-3-70B-Instruct}~\cite{dubey2024llama}, and \texttt{Command-R-Plus}~\cite{command-r-plus}.


\subsubsection{Simulated Agent Data.}
To improve the model's ability to adapt based on environmental feedback, we collect action sequences paired with observational data from various environments, represented as $\{o_0, a_1, o_1, a_2, o_2, \cdots, a_{T_g}, o_{T_g}\}$. This representation encodes the model's responses to environmental observations within its parameters. We execute official codes from agent frameworks~\cite{yao2022react, sun2024adaplanner, wang2024llms, shinn2024reflexion} in multi-step reasoning tasks (\eg, HotpotQA~\cite{yang2018hotpotqa}) and sequential decision-making tasks (\eg, ALFWorld~\cite{shridhar2021alfworld}) to collect action trajectories that involve interaction with and feedback from the environment.




\subsection{Data Quality Control Details}
\label{app:4-3}
We ensure the integrity and relevance of the collected data through continuous quality monitoring and validation procedures.
After retrieving semantically relevant data from the web corpus, we obtain a collection of noisy agent-related data. To preserve the integrity and relevance of our dataset, it is critical to continuously monitor data quality and filter out content that resembles general text rather than agent-specific data. First, we employ \texttt{Claude-3-Sonnet}~\cite{claude-3} as the data annotator to identify whether the sample belongs to agent data or a general web corpus.
Specifically, we annotate a total of $71,473$ samples from the retrieved data, identifying $37,714$ as agent-relevant and $33,767$ as general text paragraphs.
Using the annotated samples, we train a \texttt{fastText}~\cite{joulin2016fasttext} model to effectively recall additional agent-relevant web data. 
We utilize the open-source \texttt{fastText} library\footnote{\url{https://fasttext.cc}} for training, configuring the vector dimension to $256$, learning rate to $0.1$, the maximum length of word n-gram to $3$, the minimum number of word occurrences to $3$, and the number of training epochs to $3$.
After training, the \texttt{fastText} model is used to recall agent-relevant data from the remaining retrieved samples. To filter out low-quality content, we rank the collected pages based on their predicted scores from the \texttt{fastText} model and retain only the top-ranking entries. This filtering process reduces the dataset from approximately 200 billion to 80 billion tokens, ensuring that the preserved data remains highly relevant and of sufficient quality for training LLM agents.

\section{Implementation Details}
\label{app:implementation}

We use \texttt{LLaMA-3-8B}~\cite{dubey2024llama} as the backbone for our main experiments. Our training process consists of two stages. In the two-stage pre-training, we set the batch size to $512$ and train the model for $55,000$ steps in each stage, with a learning rate of $2e-4$ and weight decay of $0.01$. For the instruction fine-tuning stage, we reduce the batch size to $16$ and train the model for $24,000$ steps, using a learning rate of $1e-6$ while maintaining the same weight decay of $0.01$.
For parallel pre-training, we apply a tensor model parallel size of $8$ and a pipeline model parallel size of $2$. These values are adjusted to $4$ and $2$, respectively, for instruction fine-tuning. We use the Adam optimizer~\cite{kingma2014adam} with $\beta_1=0.9$ and $\beta_2=0.98$ for all stages. During inference, we maintain a temperature of $T=0$.
Training \texttt{\method-8B-Base} requires 128 NVIDIA A100 (40G) GPUs for 11.1 days (7.7 days for Stage I pre-training and 3.4 days for Stage II pre-training). Training \texttt{\method-8B-IFT} uses 16 NVIDIA A100 (40G) GPUs for 11.6 hours.

\section{Additional Experimental Results and Analysis}
\label{app:exp}

\subsection{Evaluation of the fastText Filter}
To evaluate the precision of the fastText classifier in filtering general text from web retrieval data, we leverage \texttt{Claude-3-Sonnet} to annotate 20K samples. We then compare the predictions from the fastText filter against these annotated ground-truth labels. The evaluation results are presented in Table~\ref{tab:filter}.
The results indicate that the fastText filter achieves an accuracy of approximately 88\%, suggesting that the filtering outcomes are reliable and trustworthy. Moreover, the higher recall score indicates that the filtered data encompasses most agent-relevant information from the retrieval.

\begin{table}[ht]
\centering
\fontsize{8}{10}\selectfont\setlength{\tabcolsep}{0.3em}
\begin{tabular}{@{}l>{}cccc}
\toprule
\textbf{Model ($\downarrow$)} & \textbf{Accuracy} & \textbf{F-1} & \textbf{Precision} & \textbf{Recall}\\\midrule
fastText & 87.46 & 87.20 & 83.42 & 91.33 \\\bottomrule
\end{tabular}
\caption{Classification results of the fastText filter.
}\label{tab:filter}
\end{table}

\subsection{Evaluation of Base Models}
As base models often struggle to follow instructions to solve problems, existing works evaluate these models using few-shot prompting~\cite{wei2022chain, shao2024deepseekmath} or by assessing the negative log-likelihood of the final answer~\cite{dubey2024llama} (e.g., selecting the correct choice). However, these evaluation methods are not suitable for agent environments for the following reasons: (1) \textbf{Task Complexity.} Agent environment tasks are significantly more complex than multiple-choice questions, requiring the generation of long sequences of actions rather than selecting a single answer. (2) \textbf{Contextual Task Requirements.} Task requirements are often intricately embedded within the context, leaving insufficient space for few-shot exemplars.
To this end, we evaluate \method-Base on three agent benchmarks (Nexus~\cite{srinivasan2023nexusraven}, API-Bank~\cite{li2023api}, and API-Bench~\cite{patil2023gorilla}) and one general benchmark (MMLU)~\cite{hendrycks2020measuring}, reporting the benchmark loss in Figure~\ref{fig:val_loss}. 




\subsection{Main Experimental Results on BFCL-v2}
\begin{table*}[h]
\centering
\fontsize{7}{9}\selectfont\setlength{\tabcolsep}{0.3em}
\begin{tabular}{@{}lcc>{}ccccccc>{}cccccc>{}c@{}}
\toprule
\textbf{Datasets ($\rightarrow$)} & \multicolumn{8}{c}{\textbf{AST}} & \multicolumn{7}{c}{\textbf{Exec}} & \textbf{BFCL-v2}\\
\cmidrule(lr){2-9} \cmidrule(lr){10-16} \cmidrule(lr){17-17}
\textbf{Models ($\downarrow$)} & \textbf{OA} & \textbf{Simple} & \textbf{Python} &  \textbf{Java} & \textbf{JS} & \textbf{MF} & \textbf{PF} & \textbf{PM} & \textbf{OA} & \textbf{Simple} & \textbf{Python} & \textbf{REST} & \textbf{MF} & \textbf{PF} &\textbf{PM} & \textbf{OA} \\\midrule
\multicolumn{16}{l}{\emph{Base LLMs}}  \\ \midrule
LLaMA-3-8B~\cite{dubey2024llama} & 0.94 & 1.3 & 1.0 & 2.0 & 1.5 & 0.5 & 0.5 & 0.5 & 0.40 & 2.0 & 1.0 & 1.0 & 0.0 & 0.0 & 0.0 & 17.77 \\
LLaMA-3.1-8B~\cite{dubey2024llama}  & 6.05 & 10.2 & 12.0 & 5.0 & 6.0 & 4.0 & 7.5 & 2.5 & 0.43 & 1.7 & 2.0 & 1.4 & 0.0 & 0.0 & 0.0 & 21.10\\
\rowcolor{teal!12} \method-8B-Base & \textbf{15.4} & \textbf{12.2} & \textbf{15.0} & \textbf{4.0} & \textbf{6.0} & \textbf{25.0} & \textbf{11.5} & \textbf{13.0} & \textbf{2.24} & \textbf{2.9} & \textbf{2.0} & \textbf{4.3} & \textbf{6.0} & \textbf{0.0} & \textbf{0.0} & \textbf{25.18}\\\midrule
\multicolumn{16}{l}{\emph{Open-Source Instruction Fine-Tuned LLMs (Small)}}  \\ \midrule
LLaMA-3-8B-Instruct~\cite{dubey2024llama} & 60.47 & 58.3 & 65.5 & 38.0	& 42.0 & 76.5 & 58.0 & \textbf{49.0} & 68.88 & 44.5 & 89.0 & 55.7 & 86.0 & \textbf{78.0} & \textbf{55.0} & 59.57\\
LLaMA-3.1-8B-Instruct~\cite{dubey2024llama} & 58.38 & 60.0 & 68.8 & 32.0 & 46.0 & 66.5 & 65.0 & 42.0 & \textbf{72.60} & 83.7 & 87.0 & 77.1 & 83.0 & 76.0 & 52.5 & 61.39\\
LLaMA-3-8B-IFT & 47.43 & 66.7 & 75.5 & 37.0 & \textbf{56.0} & 45.5 & 54.0 & 23.5 & 63.41 & \textbf{87.7} & 93.0 & \textbf{80.0} & 68.0 & 58.0 & 40.0 & 62.12\\
\rowcolor{teal!12} \method-8B-IFT & \textbf{66.39} & \textbf{72.5} & \textbf{81.8} & \textbf{45.0} & 54.0 & \textbf{79.5} & \textbf{70.5} & 43.0 & 69.82 & 85.3 & \textbf{95.0} & 71.4 & \textbf{88.0} & 66.0 & 40.0 & \textbf{70.78}\\\midrule
\multicolumn{16}{l}{\emph{For Reference: Open-Source Instruction Fine-Tuned LLMs (Medium to Large) and API-based Commercial LLMs}}  \\ \midrule
Gemini-1.5-Flash~\cite{reid2024gemini} & 77.44 & 67.3 & 92.8 & 55.0 & 54.0 & 94.0 & 71.5 & 77.0 & 73.23 & 57.9 & 93.0 & 22.9 & 86.0 & 74.0 & 75.0 & 70.75\\
Mixtral-8x22B~\cite{jiang2024mixtral} & 57.92 & 67.2 & 87.5 & 54.0 & 60.0 & 82.0 & 50.5 & 32.0 & 63.59 & 71.9 & 88.0 & 55.7 & 74.0 & 56.0 & 52.5 & 63.26\\
gpt-3.5-turbo-0125~\cite{chatgpt} & 66.31 & 63.8 & 75.3 & 50.0 & 66.0 & 78.0 & 68.0 & 55.5 & 65.88 & 44.5 & 89.0 & 0.0 & 86.0 & 78.0 & 55.0  & 66.53 \\
Claude-3-Haiku~\cite{claude-3} & 62.52 & 77.6 & 95.8 & 63.0 & 74.0 & 93.0 & 47.5 & 32.0 & 60.73 & 89.4& 96.0 & 82.9 & 94.0 & 32.0 & 27.5 & 55.47 \\
Command-R-Plus-FC~\cite{command-r-plus} & 77.65 & 69. 6 & 85.8 & 61.0 & 62.0 & 88.0 & 82.5 & 70.5 & 77.41 & 89.1 & 94.0 & 84.3 & 86.0 & 82.0 & 52.5 & 76.29\\
LLaMA-3-70B-Instruct~\cite{dubey2024llama} & 87.90 & 75.6 & 94.8 & 60.0 & 72.0 & 94.0 & 93.0 &89.0 & 88.04 & 94.1 & 94.0 & 94.3 & 94.0 & 84.0 & 80.0  & 84.95\\
gpt-4-0613~\cite{achiam2023gpt} & 91.92 & 81.2 & 95.5 & 68.0 & 80.0 & 96.0 & 96.0 & 94.5 & 87.57 & 98.3 & 98.0 & 98.6 & 96.0 & 86.0 & 70.0 & {89.26} \\\bottomrule
\end{tabular}
\caption{Main experiment results on BFCL-v2. 
}\label{tab:bfcl-v2}
\end{table*}

Table~\ref{tab:bfcl-v2} displays detailed experimental results on BFCL-v2, covering AST and Execution, two aspects in evaluation of function calling capabilities.
Aside from the notations across the other tables, ``JS'' indicates ``JavaScript''; ``MF'', ``PF'', and ``PM'' refer to ``multiple functions'', ``parallel functions'', ``parallel multiple functions''.
The superior performance of \texttt{\method-3-8B} in AST evaluations indicates that the pre-training stage successfully introduced syntax knowledge of function calling into the model, which also contributes to improvements in the Execution aspect. However, the performance gain in Execution evaluations is less pronounced. This is because, lacking access to the instruction fine-tuning data used for \texttt{LLaMA-3-8B}, our \texttt{\method-8B-IFT} demonstrates limited instruction-following capabilities compared to \texttt{LLaMA-3-8B-Instruct} and \texttt{LLaMA-3.1-8B-Instruct}. Consequently, it is more challenging to follow instructions to generate executable functions.

\subsection{Effect of Backbone LLMs}
\begin{table}[ht]
\centering
\fontsize{8}{10}\selectfont\setlength{\tabcolsep}{0.4em}
\begin{tabular}{@{}l>{}cccccc>{}c@{}}
\toprule
\textbf{Datasets ($\rightarrow$)} & \multicolumn{7}{c}{\textbf{AgentBench}} \\
\cmidrule(lr){2-8}
\textbf{Models ($\downarrow$)} & \textbf{OA} & \textbf{OS} & \textbf{DB} &  \textbf{HH} & \textbf{KG} & \textbf{WB} & \textbf{WS} \\\midrule
Mistral-7B-v0.3-Base~\cite{jiang2023mistral} & 0.40 & 7.6	& 0.7	& 0.0 & 8.9	& 11.0 &	1.4 \\
\rowcolor{teal!12} \method-7B-Base (Mistral) & \textbf{1.46} & \textbf{18.3} & \textbf{21.0} & \textbf{24.0} & \textbf{12.7} & \textbf{14.0} & \textbf{46.2}\\\midrule
Mistral-7B-v0.3-Instruct~\cite{jiang2023mistral} & 1.10 & 18.1 & 15.0	& 4.0 & 	8.9	& 18.0	& 39.6 \\
Mistral-7B-v0.3-IFT & 1.32	& 17.4	& \textbf{18.0}	& 8.0	& 15.9	& 20.0	& 45.1\\
\rowcolor{teal!12} \method-7B-IFT (Mistral) & \textbf{1.72} & \textbf{17.4} & 11.7 & \textbf{30.0} & \textbf{20.1} & \textbf{25.0} & \textbf{55.4}\\\bottomrule
\end{tabular}
\caption{Experimental results of \method-7B (Mistral) with \texttt{Mistral-7B-v0.3} as backbone LLM on AgentBench.
}\label{tab:mistral}
\end{table}

Table~\ref{tab:mistral} reports the performance of \method and the baselines using \texttt{Mistral-7B-v0.3} as backbone LLM on AgentBench. 
Notably, there exist consistent gains in terms of the average performance on both base model and instruction-tuned model ($1.06$ on base model and $0.4$ on IFT model), justifying the advantage of pre-training on \dataset across different LLM types and architectures.

\section{Case Studies}
\label{app:case}

\subsection{Code-to-Text Synthesis Example}

We present an example of synthesized API documentation as follows:
\vspace{1ex}
\begin{Verbatim}
**API: `easyai.model.base_block.utility.utility_block.ConvBNActivationBlock`**

The `ConvBNActivationBlock` is a utility class that combines a convolutional layer, a batch 
normalization layer, and an activation layer into a single block. This block is commonly used as a 
building block in various neural network architectures, including the PNASNet architecture 
demonstrated in the provided code.

**Parameters:**

- `in_channels` (int): The number of input channels for the convolutional layer.
- `out_channels` (int): The number of output channels for the convolutional layer.
- `kernel_size` (int or tuple): The size of the convolutional kernel.
- `stride` (int or tuple, optional): The stride of the convolutional operation. Default is 1.
- `padding` (int or tuple, optional): The padding added to the input tensor before applying 
    the convolution. Default is 0.
- `bias` (bool, optional): Whether to include a bias term in the convolutional layer. Default is 
    `False`.
- `bnName` (str or `easyai.base_name.block_name.NormalizationType`, optional): The normalization 
    type to be used. Default is `NormalizationType.BatchNormalize2d`.
- `activationName` (str or `easyai.base_name.block_name.ActivationType`, optional): The activation 
    function to be used. Default is `ActivationType.ReLU`.

**Attributes:**

- `conv`: The convolutional layer.
- `bn`: The batch normalization layer.
- `activation`: The activation layer.

**Methods:**

- `forward(x)`: Defines the forward pass of the block. It takes an input tensor `x` and applies the 
    convolutional, batch normalization, and activation operations sequentially.
- `get_name()`: Returns the name of the block, which is a combination of the class name and a unique 
    index.

**Example Usage:**

```python
import torch
from easyai.model.base_block.utility.utility_block import ConvBNActivationBlock

# Create a ConvBNActivationBlock with input channels 3, output channels 64, kernel size 3x3, and 
    stride 1
block = ConvBNActivationBlock(in_channels=3, out_channels=64, kernel_size=3, stride=1)

# Generate a random input tensor
x = torch.randn(1, 3, 32, 32)

# Pass the input tensor through the block
output = block(x)
```

In the provided code, the `ConvBNActivationBlock` is used as the first layer of the PNASNet 
architecture, where it takes the input image data and applies a convolutional operation followed by 
batch normalization and activation.
\end{Verbatim}

\subsection{Retrieved Data Examples}

We present two examples of high-quality retrieved data as follows:

\vspace{1ex}
\begin{Verbatim}
The Cardboard Kitchen : 6 Steps
By nicholasniski01 in Craft Cardboard
Introduction: The Cardboard Kitchen
In this instructable I will show you how to make a Cardboard Kitchen. The Cardboard Kitchen Is 
    almost entirely made out of Cardboard. This Kitchen includes a Stove, Oven, Sink, Dishwasher, 
    Fridge and Microwave.
lots of small boxes
and a medium size box
Step 1: How to Make a Fridge
    you need to disassemble the medium size box(Get rid of ALL the tape)...
Step 2: How to Make a Microwave
    First get a small box, make a rectangular hole in the box...
Step 3: How to Make a Sink
    First get a small box, cut off the top of the box...
Step 4: How to Make a Dishwasher
    Cut out a square of cardboard for the size of the dishwasher then you color the cardboard black, 
    silver or any other color you would want for the dishwasher...
Step 5: How to Make a Stove
    To make the stove you make a black circle with for lines going out of the circle on an unused 
    section of your big box or \"counter\"...
Step 6: How to Make an Oven
    Get a square the size you want your oven to be...
\end{Verbatim}

\vspace{1ex}
\begin{Verbatim}
manual Prestigio MultiReader 5574
You can create your event and make a plan on your calendar. On the home screen or list menu, 
    tap Calendar. View the calendar On the home screen or list menu, tap Calendar to check the 
    calendar. Tap to change your calendar to Day, Week,Month or Agenda view. Create an event 
1. Go to Calendar, select a date. 
2. Tap to create a new event. 
3. Edit reminder settings. 
4. Tap Done to save the event. 
\end{Verbatim}

\subsection{Data Quality Filtering Failure Cases}

%
We present a failure case of the fastText filter below:
\vspace{1ex}
\begin{Verbatim}
[TEXT]
We're making it even easier for you to stay connected to 99ROCK wherever you go! Besides tuning 
    in on your radio, you can also stream your favorite station through your computer, smartphone, 
    tablet, and your smart speaker.
If you are in or near the Fort Walton Beach-Destin broadcast area, tune your radio to: 99.5 FM
Stream 99ROCK at work or home from your computer on one of these web players: Triton Player iHeart 
    Radio TuneIn
Listen to 99ROCK on-the-go thru one of these popular streaming apps or thru the 99ROCK mobile 
    app: iOS App Google Play iHeart Radio TuneIn
First you need to enable the 99ROCK skill:
Say, ``Alexa, enable the ninety-nine rock Skill''
After you have enabled the Skill, listen to our station just by saying "Alexa, open ninety nine 
    rock"
Just say, ``Hey Google, play ninety-nine rock''
[CATEGORY]
Agent
\end{Verbatim}
In this case, the fastText model incorrectly categorized the text as agent-relevant data. This misclassification likely occurred because fastText relies on gram frequency analysis, and the presence of multiple high-tech terms (e.g., iOS, App, Google Play) in the paragraph may have misled the model.
\section{Prompt Templates}
\label{app:prompt}

\subsection{Prompt Template for Code-to-Text Synthesis}
\vspace{2ex}
\begin{Verbatim}
Please use your knowledge to write an API documentation for the given APIs and consider the given 
code as the example usage.

API:
{api}

Code:
{code}

API Documentation:
\end{Verbatim}

\subsection{Prompt Template for LLM Annotator in Data Quality Control}
\vspace{2ex}
\begin{Verbatim}
Please categorize the given text belong to agent-relevant data or other general text. The 
    definitions are as follows:
1. Agent: Tool documentation text that describes the usage of a tool, software, or API; and action 
    trajectory text that describes a sequence of actions or steps to achieve a goal.
2. General: Other general text that does not belong to the above two categories.
Below are some examples:
[TEXT]
**API: `easyai.model.base_block.utility.utility_block.ConvBNActivationBlock`**  
The `ConvBNActivationBlock` is a utility class that combines a convolutional layer, a batch 
normalization layer, and an activation layer into a single block. This block is commonly used as a 
building block in various neural network architectures, including the PNASNet architecture 
demonstrated in the provided code.  **Parameters:**  - `in_channels` (int): The number of input 
channels for the convolutional layer. 
- `out_channels` (int): The number of output channels for the convolutional layer. 
- `kernel_size` (int or tuple): The size of the convolutional kernel. 
- `stride` (int or tuple, optional): The stride of the convolutional operation. Default is 1. 
- `padding` (int or tuple, optional): The padding added to the input tensor before applying the 
    convolution. Default is 0.
- `bias` (bool, optional): Whether to include a bias term in the convolutional layer. Default is 
    `False`. 
- `bnName` (str or `easyai.base_name.block_name.NormalizationType`, optional): The normalization 
    type to be used. Default is `NormalizationType.BatchNormalize2d`. 
- `activationName` (str or `easyai.base_name.block_name.ActivationType`, optional): The activation 
    function to be used. Default is `ActivationType.ReLU`.
[CATEGORY]
Agent

[TEXT]
I want to deliver a Birthday Gift to my friend in London, UK. Then, I need to book a flight from 
New York, USA to London, UK on August 1st, 2023 for myself. After arriving in London, I would like 
to see Dr. Smith for my Migraine. Once my health is in check, I'd like to apply for a Software 
Engineer job in London. 
Step 1: Call deliver_package API with package: 'Birthday Gift' and destination: 'London, UK' 
deliver_package(package=Birthday Gift, destination=London, UK) 
Step 2: Call book_flight API with date: '2023-08-01', from: 'New York, USA' and to: 'London, UK' 
book_flight(date=2023-08-01, from=New York, USA, to=London, UK) 
Step 3: Call see_doctor_online API with disease: 'Migraine' and doctor: 'Dr. Smith' 
see_doctor_online(disease=Migraine, doctor=Dr. Smith) 
Step 4: Call apply_for_job API with job: 'Software Engineer' apply_for_job(job=Software Engineer)"
[CATEGORY]
Agent

[TEXT]
My New Crock Pot -- Creuzer Leave a comment on My New Crock Pot I went out and got myself a new 
crock pot. I rather like this one. It has 3 settings, High, Low, and Warm. It is designed to 
be hauled around even! There are latches on each side of the lid to clip the lid in place. It 
even came with it's own spoon that clips into the lid! A really neat feature is that the lid has 
little tabs so you can set the lid on one of the handles and it won't go all sliding all over 
the place. I think this is a winner, I plan on using it a lot for the cooking club I am in.
[CATEGORY]
General

Please categorize the following text into agent-relevant data (Agent) or general text (General).
ONLY respond the category name (Agent/General) for each text. If you are unsure, please respond with 
    'General'.
[TEXT]
{text}
[CATEGORY]
\end{Verbatim}

\begin{table*}[ht]
\centering 
\renewcommand\arraystretch{0.98}
\fontsize{8}{10}\selectfont \setlength{\tabcolsep}{0.4em}
\begin{tabular}{@{}lcccp{0.4\linewidth}@{}}
\toprule
\textbf{Data Source} & \textbf{Type} & \textbf{Format} & \textbf{Tokens (B)} & \textbf{URL Link}\\\midrule 
ToolBench~\cite{qin2023toolllm} & Traj. & Dialog & 0.530 & \url{https://github.com/OpenBMB/ToolBench}\\
AgentInstruct~\cite{zeng2023agenttuning} & Traj. & ReAct & 0.002 & \url{https://huggingface.co/datasets/THUDM/AgentInstruct}\\
Alexa-Arena~\cite{gao2024alexa} & Traj. & NL Plan & 0.035 & \url{https://github.com/amazon-science/alexa-arena/tree/main} \\
chat\_ego\_4d~\cite{mu2024embodiedgpt} & Traj. & API Seq & 0.025 & \url{https://github.com/EmbodiedGPT/EgoCOT_Dataset}\\
FireAct~\cite{chen2023fireact} & Traj. & ReAct & 0.002 & \url{https://fireact-agent.github.io/}\\
NAT~\cite{wang2024learning} & Traj. & ReAct & 0.003 & \url{https://github.com/Reason-Wang/NAT}\\
ToolAlpaca~\cite{tang2023toolalpaca} & Traj. & Plain Text & 0.004 & \url{https://github.com/tangqiaoyu/ToolAlpaca/tree/main}\\
Lumos~\cite{yin2024agent} & Traj. & Dialog & 0.109 & \url{https://huggingface.co/datasets/ai2lumos/lumos_complex_qa_ground_iterative?row=0} \\
STE~\cite{wang2024llms} & Traj. & Plain Text & 0.025 & \url{https://github.com/microsoft/simulated-trial-and-error} \\
toolbench~\cite{xu2023on} & Traj. & API Seq & 0.010 & \url{https://github.com/sambanova/toolbench}\\
Gorilla~\cite{patil2023gorilla} & Doc. & API Seq & 0.009 & \url{https://gorilla.cs.berkeley.edu/}\\
PublicAPIs & Doc. & Plain Text & 0.008 & \url{https://github.com/public-apis/public-apis?tab=readme-ov-file} \\
TaskBench~\cite{shen2023taskbench} & Traj. & NL Plan & 0.020 & \url{https://github.com/microsoft/JARVIS/tree/main/taskbench} \\
RestBench~\cite{song2023restgpt} & Traj. & API Seq & 0.001 & \url{https://github.com/Yifan-Song793/RestGPT/tree/main/datasets} \\
SayCanPay~\cite{hazra2024saycanpay} & Traj. & NL Plan & 0.001 & \url{https://github.com/RishiHazra/saycanpay} \\
AgentFlan~\cite{chen2024agent} & Traj. & Dialog & 0.020 & \url{https://github.com/InternLM/Agent-FLAN} \\
PlanBench~\cite{valmeekam2024planbench} & Traj. & NL Plan & 0.001 & \url{https://github.com/karthikv792/LLMs-Planning} \\
SwftSage~\cite{lin2024swiftsage} & Traj. & NL Plan & 0.022 & \url{https://github.com/yuchenlin/SwiftSage} \\
T-Eval~\cite{chen2023t} & Traj. & Dialog & 0.040 & \url{https://github.com/open-compass/T-Eval} \\
API-Bank~\cite{li2023api} & Traj. & API Seq & 0.001 & \url{https://github.com/AlibabaResearch/DAMO-ConvAI/tree/main/api-bank} \\
JeirchoWorld~\cite{ammanabrolu2021modeling} & Traj. & NL Plan & 0.001 & \url{https://github.com/JerichoWorld/JerichoWorld} \\
API-Pack~\cite{guo2024api} & Traj. & API Seq & 0.800 & \url{https://huggingface.co/datasets/zguo0525/API-Pack/tree/main}\\
CodeAct~\cite{lv2024codeact} & Traj. & Dialog & 0.009 & \url{https://huggingface.co/datasets/xingyaoww/code-act} \\
UltraTool~\cite{huang2024planning} & Traj. & NL Plan & 0.002 &\url{https://github.com/JoeYing1019/UltraTool/tree/main} \\
Tooleyes~\cite{ye2024tooleyes} & Doc. & JSON & 0.001 & \url{https://github.com/Junjie-Ye/ToolEyes/tree/main}\\
OpenMathInstruct~\cite{toshniwal2024openmathinstruct} & Traj. & API Seq & 0.335 & \url{https://huggingface.co/datasets/nvidia/OpenMathInstruct-1}\\
NexasRaven~\cite{srinivasan2023nexusraven} & Traj. & JSON & 0.001 & \url{https://huggingface.co/Nexusflow}\\
Seal-Tools~\cite{wu2024seal} & Traj. & API Seq & 0.002 & \url{https://github.com/fairyshine/Seal-Tools/tree/master} \\
UltraInteract~\cite{yuan2024advancing} & Traj. & QA & 0.16 & \url{https://huggingface.co/datasets/openbmb/UltraInteract_sft?row=0} \\
Python Module & Doc. & Plain Text & 0.001 & \url{https://docs.python.org/3.12/} \\
AgentTraj-L~\cite{xi2024agentgym} & Traj. & Dialog & 0.020 & \url{https://huggingface.co/datasets/AgentGym/AgentTraj-L} \\
MNMs~\cite{ma2024mms} & Traj. & API Seq & 0.001 & \url{https://huggingface.co/datasets/zixianma/mnms} \\
PythonQA-API-Usage & Doc. & QA & 0.003 & \url{https://huggingface.co/datasets/RazinAleks/SO-Python_QA-API_USAGE_class} \\
APIText & Traj. & API Seq & 0.001 & \url{https://huggingface.co/datasets/havens2/apitext} \\
StarCoder-APIs~\cite{lozhkov2024starcoder} & Traj. & Code & 6.147 & \url{https://huggingface.co/datasets/luna-code/starcoderdata-apis} \\
APIs\_v2 & Traj. & API Seq & 0.003 & \url{https://huggingface.co/datasets/vinilazzari/apis_v2} \\
Ultimate & Traj. & QA & 0.002 & \url{https://huggingface.co/datasets/Kris8an/ultimate_apicalls_and_topbot}\\
xLAM~\cite{zhang2024xlam} & Doc. & QA & 0.022 & \url{https://huggingface.co/datasets/Salesforce/xlam-function-calling-60k} \\\bottomrule
\end{tabular}
\caption{Data sources of the seed data in \dataset. }
\label{tab:data-source1}
\end{table*}

\begin{table*}[ht]
\centering 
\renewcommand\arraystretch{0.98}
\fontsize{8}{10}\selectfont \setlength{\tabcolsep}{0.4em}
\begin{tabular}{@{}lcccp{0.45\linewidth}@{}}
\toprule
\textbf{Data Source} & \textbf{Type} & \textbf{Format} & \textbf{Tokens (B)} & \textbf{URL Link}\\\midrule 
API\_doc & Doc. & Plain Text & 0.001 & \url{https://huggingface.co/datasets/Prakhar1000/API_Documentation_dataset_alpaanco?row=0} \\
ChatsBug & Traj. & NL Plan & 0.009 & \url{https://huggingface.co/datasets/chats-bug/agent_action_plan?row=0} \\
sample\_scripts & Traj. & API Seq & 0.002 & \url{https://huggingface.co/datasets/prantadi/tokenized_dataset_1024_SampleScripts_deduped_API-ref?row=1} \\
Agent-Trajectories & Traj. & API Seq & 0.001 & \url{https://huggingface.co/datasets/Agent-Eval-Refine/Agent-Trajectories/tree/main} \\
Agent-Instruct & Traj. & Dialog & 0.056 & \url{https://huggingface.co/datasets/sam-mosaic/agent-instruct} \\
Agent007 & Traj. & API Seq & 0.001 & \url{https://huggingface.co/datasets/DepositorOP/agent007} \\
AgentCode & Traj. & API Seq & 0.010 & \url{https://huggingface.co/datasets/AlignmentLab-AI/agentcode} \\
syn-web-agent & Traj. & JSON & 0.001 & \url{https://huggingface.co/datasets/allyson-ai/synthetic-web-agent} \\
syn-llama & Traj. & Dialog & 0.004 & \url{https://huggingface.co/datasets/Cyleux/agent-machine-convo-llama-nicholas-2k-gpt4-verified} \\
seq-Mind2Web & Traj. & JSON & 1.243 & \url{https://huggingface.co/datasets/Izazk/Sequence-of-action-prediction-mind2web} \\
syn-gemma & Traj. & Dialog & 0.047 & \url{https://huggingface.co/datasets/NickyNicky/function-calling-sharegpt_chatml_gemma_agent} \\
LLM Robot & Traj. & API Seq & 0.001 & \url{https://huggingface.co/datasets/Aryaduta/llm_robot} \\
Verifiers for Code & Traj. & Plain Text & 0.05 & \url{https://huggingface.co/datasets/verifiers-for-code/CodeNet-Planner} \\ 
isotonic planner & Traj. & NL Plan & 0.005 & \url{https://huggingface.co/datasets/Isotonic/planner_dataset} \\ 
Turing Solutions & Traj. & NL Plab & 0.001 & \url{https://huggingface.co/datasets/TuringsSolutions/GlobalFunctionCallingTrainingSetLarge} \\
G-PlanET & Traj.  & NL Plan & 0.003 & \url{https://huggingface.co/datasets/TuringsSolutions/GlobalFunctionCallingTrainingSetLarge} \\
Pandas Doc & Doc. & Plain Text & 0.004 & \url{https://pandas.pydata.org/} \\
Sugarcrm & Doc. & Plain Text & 0.001 & \url{https://huggingface.co/datasets/kaahila/sugarcrm_130_documentation} \\
AWS & Doc. & Plain Text & 0.033 & \url{https://huggingface.co/datasets/sauravjoshi23/aws-documentation-chunked}  \\
LangChain & Doc. & Plain Text & 0.005 & \url{https://huggingface.co/datasets/jamescalam/langchain-docs-23-06-27} \\
Code Library & Doc. & Plain Text & 0.013 & \url{https://huggingface.co/datasets/code-rag-bench/library-documentation}\\
PublicAPIs-extend & Doc. & Plain Text & 0.718 & \url{https://github.com/public-apis/public-apis?tab=readme-ov-file} \\
Torch & Doc. & Plain Text & 0.005 & \url{https://pytorch.org/docs/stable/index.html} \\\bottomrule
\end{tabular}
\caption{Data sources of the seed data in \dataset (Cont'd). }
\label{tab:data-source2}
\end{table*}

\end{document}